\documentclass{article}
\usepackage{graphicx}
\usepackage{amsmath}
\usepackage{float}
\usepackage{subcaption}

\PassOptionsToPackage{numbers, compress}{natbib}

\usepackage[preprint]{neurips_2026}

\usepackage[utf8]{inputenc} % allow utf-8 input
\usepackage[T1]{fontenc}    % use 8-bit T1 fonts
\usepackage{hyperref}       % hyperlinks
\usepackage{url}            % simple URL typesetting
\usepackage{booktabs}       % professional-quality tables
\usepackage{makecell}       % multi-line cells for table headers
\usepackage{tabularx}       % tabularx for fixed-width auto-wrapping columns
\usepackage{amsfonts}       % blackboard math symbols
\usepackage{nicefrac}       % compact symbols for 1/2, etc.
\usepackage{microtype}      % microtypography
\usepackage{xcolor}         % colors
\usepackage[most]{tcolorbox}
\tcbuselibrary{listings}
\newtcblisting{promptbox}[1][]{
  listing only,
  listing options={basicstyle=\ttfamily\small, breaklines=true,
    breakatwhitespace=true, columns=fullflexible, upquote=true,
    keepspaces=true, showstringspaces=false},
  colback=gray!5, colframe=black!60, fonttitle=\bfseries,
  breakable, enhanced, sharp corners, boxrule=0.5pt,
  left=4pt, right=4pt, top=4pt, bottom=4pt, #1
}
\newcommand{\myparagraph}[1]{\vspace{0pt} \noindent \textbf{#1} \hspace{3pt}}

\title{Vision-Language Binding in In-Context Image Generation}

\author{
Chris Ge$^{1}$\thanks{Correspondence to \texttt{cge7@mit.edu}} \quad Rohit Gandikota$^{2}$ \quad \textbf{Antonio Torralba}$^1$ \quad \textbf{Tamar Rott Shaham}$^{1}$\\
\\
$^1$MIT CSAIL \quad $^2$Northeastern University
}

\begin{document}

\maketitle

\begin{abstract}
In-context image generation models such as FLUX.2 \citep{flux-2-2025} take a text prompt and an optional reference image as visual conditioning for the output. Internally, all three inputs --- text, reference image, and the noise tokens --- are concatenated and processed through a single attention stream, where all tokens can attend to one another. This leaves open how reference information flows through the model to produce the output image. We show that an implicit cross-modal binding emerges between the text tokens and the reference image: the text tokens absorb visual reference content during the forward pass, and that absorbed content causally influences the generated output. We surface this binding with three causal interventions on FLUX.2: T2I Lens, which decodes intermediate text-token activations through a text-to-image path; Attention Knockout, which severs specific attention edges; and I2I-to-I2I Patching, which copies text token activations between editing runs. Across 2,875 editing tasks on various images, including SUN397 and DreamBench++ datasets and images collected online, we observe a consistent division of labor: properties of the reference image, like color, style, and scene setting, are first written into the text tokens, which carry them to the generated image; pixel-exact properties like a specific face or instance identity bypass the text tokens and flow directly from reference to image through image-to-image attention. We further localize the reference-text binding to the padding tokens of the text sequence. These results show that text tokens in a multimodal DiT are not just prompt holders, but a structured channel for reference image content. More broadly, they suggest that even in unified-attention multimodal generative models, token modality structures how conditioning information is represented and routed across the network.\footnote{Our code and data are available at \url{https://chrisg777.github.io/i2i-interp/}.}

\end{abstract}
\section{Introduction}
\label{sec:Introduction}

A single FLUX.2 forward pass can render a scene from a prompt, edit a reference image in response to an instruction, or stage a reference object into a completely new context. The same set of weights is used for all three. How does this happen, and what information is actually flowing between text tokens, reference image tokens, and generated image tokens inside the model when it does?

In earlier image-to-image (I2I)  diffusion models, this question does not arise by construction. Methods like Prompt-to-Prompt \citep{hertzprompt} and InstructPix2Pix \citep{brooks2023instructpix2pix} layered editing on top of a text-to-image U-Net \citep{ronneberger2015u}, with the reference image entering through a fixed conditioning channel and the instruction coming in through cross-attention. Text and image had distinct architectural roles, and their interaction was confined to specific layers one could point to directly.

OmniGen \citep{xiao2025omnigen} and later FLUX.1 Kontext \citep{labs2025flux} and FLUX.2 \citep{flux-2-2025} changed this paradigm. Architecturally, they routed reference-image tokens into the same multimodal attention sequence as the text tokens, with no dedicated cross-attention layer for reference conditioning. They were also trained to also handle text-to-image (T2I)  generation natively, in a single set of weights. The same multimodal attention block now switches, on a per-input basis, between a T2I forward pass with no reference  and a reference-conditioned forward pass where reference content sits in the input sequence.

This raises a question the architecture itself doesn't answer: what role do the text tokens play when a reference image is included in the input? Are they still just prompt holders, as their text-to-image training would suggest, or do they pick up reference image content during the forward pass and serve as a medium that carries it into the generated image? An analogous question was recently posed for vision-language models that generate text from image plus text \citep{kaduri2025s}; we take it up for diffusion transformers that generate images from image plus text.

\begin{figure}[t]
    \centering
    \includegraphics[width=1.0\linewidth]{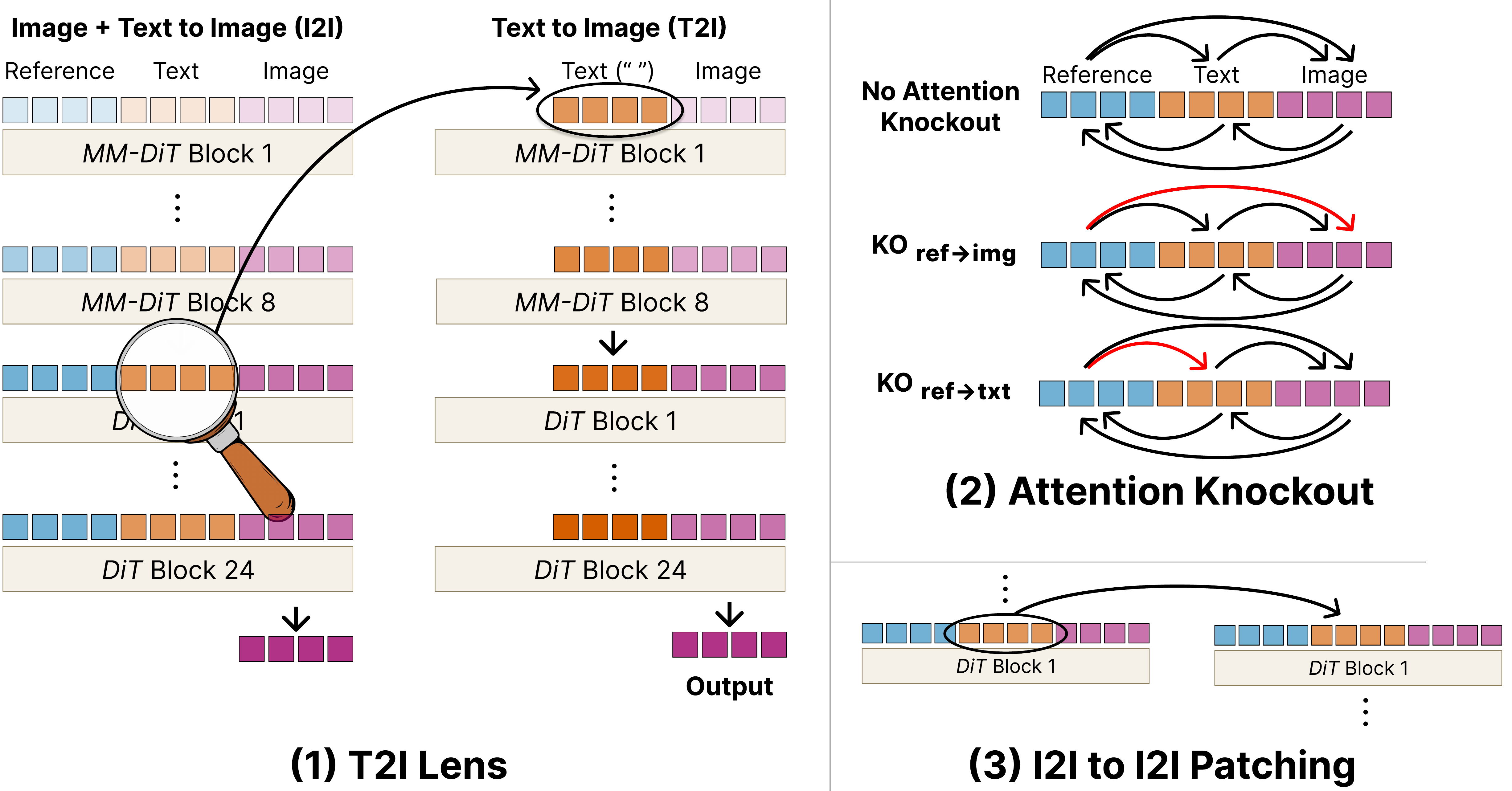}
    \caption{\textbf{Our three causal intervention methods on text tokens.} (1) In \textit{T2I Lens}, we patch the text residual stream after a given MM-DiT or DiT block into the raw text embeddings of a reference-free, unconditional text-to-image generation with an \textit{empty prompt}, decoding the text-token activations at that point back into pixel space. (2) In \textit{Attention Knockout} \citep{geva2023dissecting}, we mask out the attention paid by text tokens to reference tokens, or by image tokens to reference tokens, closing one possible pathway for reference information and isolating its causal effect. (3) In \textit{I2I-to-I2I Patching}, we patch the same intermediate-layer text activations as T2I Lens from a source image-editing generation into a target image-editing generation that shares the same instruction prompt, testing whether the absorbed reference information transfers to the patched run's output.}
    \label{fig:methods}
\end{figure}

We answer these questions with three causal interventions on the multimodal attention stream of FLUX.2, illustrated in Figure \ref{fig:methods}. \textit{T2I Lens} decodes the intermediate-layer text-token activations of a reference-conditioned forward pass through a text-to-image path, to read out what the text tokens encode at that point. \textit{Attention Knockout} blocks specific attention edges, to test which pathways the model uses for each kind of reference content. \textit{I2I-to-I2I Patching} copies text-token activations from one editing run into another to test which properties causally transfer with them.

We find that the text tokens carry significantly more information about the reference image than their original training role as prompt holders would suggest. What they carry follows a clear pattern: properties that can be expressed in words are first written into the text tokens, which carry them to the generated image, while pixel-exact properties bypass the text tokens entirely and flow directly from reference to image. We further localize this binding to a specific region of the text sequence that has not, to our knowledge, been studied as a meaningful site of computation in this kind of model.

Beyond FLUX.2, these findings have implications for unified-attention multimodal generative models more generally. A shared attention stream does not erase the modality of its tokens: each set continues to carry the kind of conditioning content best suited to its modality, even though no part of the training objective explicitly enforces this. The diagnostic tools we develop apply to any in-context multimodal model that mixes text and image tokens in a shared attention operation, and the routing patterns we identify can inform future design choices about which modality should carry which kind of conditioning.

\section{Related Work}
\label{sec:related_works}

\subsection{Diffusion and Flow-Matching Transformers}

In FLUX.2, text tokens and image tokens share a unified attention operation through  multimodal diffusion transformer (MM-DiT) blocks \cite{esser2024scaling}.  Concretely, FLUX.2 stacks \textit{double stream blocks}, which are traditional MM-DiT blocks that maintain modality-specific QKV and MLP weights, followed by \textit{single stream blocks}, which are closer to the original DiT \citep{peebles2023scalable} in applying a single shared set of weights to the combined token sequence. In both cases, there is no privileged cross-attention path between modalities, which is what makes the routing question we ask non-trivial: there is no single layer one can point at as the place where text and image meet. 

\subsection{In-Context Image Editing}
\label{subsec:image_editing}

Beyond the methods discussed in the introduction, several DiT-based editors are relevant to our setup. ICEdit \citep{zhang2026enabling} and Group Relative Attention Guidance \citep{zhang2025group} both operate on the unified MM-attention stream, via diptych inversion and attention-bias extraction respectively, providing evidence that the attention block itself encodes structured editing behavior.  % ICEdit performs in-context editing by inverting a reference image into a diptych and reconstructing one half while editing the other, exploiting the same kind of unified attention stream we study. Group Relative Attention Guidance \citep{zhang2025group} extracts an editing bias from the shared query and key vectors of MM-attention to steer edits without retraining, providing further evidence that the attention block itself encodes structured editing behavior. 
Qwen-Image-Edit \citep{wu2025qwenimagetechnicalreport} takes a deliberately different architectural route, feeding the reference image through a Qwen2.5-VL encoder for semantic control and through a separate VAE encoder for visual appearance, splitting reference content into a ``what is in it'' path through the language stream and a ``how exactly does it look'' path through the latent image stream. In retrospect, this explicit split anticipates the routing asymmetry we find inside FLUX.2's single attention stream.

\subsection{Interpretability of Multimodal Generative Models}

Prior interpretability work has focused on the no-reference, T2I setting. Diffusion Lens \citep{toker2024diffusion} generates images from   intermediate text-encoder activations to reveal  concept representations; we apply  this decoding strategy to the text tokens inside a MM-DiT after they have already attended to a reference image. ConceptAttention \citep{helblingconceptattention} showed that DiT attention layers learn highly interpretable concept-localized features, implying the attention representations carry recoverable conceptual structure. Causal mediation analysis has been applied to T2I to localize visual attribute knowledge in the CLIP text encoder \citep{basu2024localizing}. Other work zooms in on the structure of the text stream itself: \citet{toker2025padding} showed that padding tokens take on a register-like role that influences generation, and \citet{kaplan2025follow} traced how information distributes across textual tokens in the final layers of T2I models. Our padding-token observation extends this line to a setting where padding absorbs information from a second modality rather than from the text encoder alone. I2AM \citep{park2025i2am} also studies I2I interpretability, but differs both architecturally (UNet-based latent diffusion vs.\ MM-DiT) and in task scope (spatially-constrained settings like virtual try-on and inpainting vs.\ free-form in-context editing). 

The closest analogue to our question comes from the vision-language model (VLM) side. \citet{kaduri2025s} found that the textual query tokens of a VLM store global image information, while fine-grained visual attributes flow directly from spatially localized image regions. We ask the diffusion-transformer counterpart of their question, and use causal attention interventions in the style of \citet{geva2023dissecting} and \citet{heimersheim2024use} to answer it. To our knowledge, no prior work has used such interventions on the multimodal attention stream of an in-context image editor to ask which modality carries which kind of reference information.
\section{Methods}
\label{sec:methods}

We probe the role of text tokens during reference-conditioned editing in FLUX.2 with three causal interventions on its multimodal attention stream. Throughout the paper, we use three labels: \textbf{Reference} tokens are the tokens of the reference image fed into the image-to-image (\textbf{I2I}) edit, \textbf{Text} tokens are the tokens of the edit instruction, and \textbf{Image} tokens are the noisy latents that become the generated output. Figure~\ref{fig:methods} gives an overview of the three methods.

\textbf{T2I Lens.} \textit{Asks what reference content the text tokens have absorbed during a reference-conditioned forward pass.} We run a normal I2I edit and grab the text-token activations at one chosen intermediate layer. We then start a separate text-to-image (\textbf{T2I}) generation that has no reference image and an empty text prompt (which still produces padding text tokens), and we copy the saved activations onto its raw text embeddings (the input to MM-DiT layer 0). We copy the same saved activations into all $4$ denoising steps, since the text activations are otherwise reset at every step. The image that this T2I run produces is what we call the T2I Lens output. Because the T2I run has no reference and no instruction of its own, anything in its output that resembles the original reference must have ridden in on the patched activations. In practice we find that generic reference content shows up after the $8$th double stream block in the first denoising step (Appendix~\ref{sec:mm7_single9}). Consistently patching at the same layer across different tasks lets us point at a consistent site in the computation where the vision-language binding has happened. For color transfer tasks we use a small variant of T2I Lens, patching at the $10$th single-stream block with a single denoising step, which surfaces the reference color more cleanly (Appendix~\ref{subsec:t2i_color}).

\textbf{Attention Knockout} \citep{geva2023dissecting}. \textit{Asks which attention pathway each kind of reference content uses to reach the generated image.} We prevent one set of tokens from attending to another at every attention layer in the model. We use this to block image tokens from attending to the reference ($\text{KO}_{\text{ref}\rightarrow\text{image}}$), or to block text tokens from attending to the reference ($\text{KO}_{\text{ref}\rightarrow\text{text}}$). Each cut closes off one of the two pathways by which reference content could reach the generated image. We run one additional, stronger knockout: up to a chosen cutoff layer we apply $\text{KO}_{\text{ref}\rightarrow\text{image}}$, and past the cutoff we block all attention into and out of the reference ($\text{KO}_{\text{ref}\rightarrow\text{image}}$, $\text{KO}_{\text{ref}\rightarrow\text{text}}$, $\text{KO}_{\text{image}\rightarrow\text{ref}}$, and $\text{KO}_{\text{text}\rightarrow\text{ref}}$). This isolates the hypothesized pathway, leaving the text tokens as the reference's only route to the output, and tests whether the reference's color and style still reach it (Appendix~\ref{sec:early_ref_drop}).

\textbf{I2I-to-I2I Patching.} \textit{Asks which reference properties causally transfer with the text-token activations alone.} We run two I2I edits side-by-side, a source and a target, that share the same edit instruction but use different reference images. We then copy the text-token activations from the source run into the target run at the same layer as in T2I Lens, and at every denoising step. If the text tokens causally carry a property of the source reference, the target output should exhibit that property.

\paragraph{Evaluation.} For each experiment we use a VLM-as-judge \cite{zheng2023judging} to give a yes/no verdict on whether the intervention transferred or removed the property we care about. The judge is \texttt{Claude-Opus-4.7}~\cite{anthropic2026opus47card}. We pass it the reference image, any I2I baselines, the intervention output, and a task-dependent question; the judging prompts are in Appendix~\ref{sec:vlm-judge-prompts}. We report accuracy pooled over all judgments in a given task family. Each entry in Tables~\ref{tab:t2i_lens}, \ref{tab:attention_knockout}, \ref{tab:i2i_to_i2i}, and \ref{tab:attention_knockout_full} is reported as $\bar{p}_{-\delta_{\text{lo}}}^{+\delta_{\text{hi}}}$, where $\bar{p}$ is the success rate over all binary judgments in the cell and $[\bar{p}-\delta_{\text{lo}},\, \bar{p}+\delta_{\text{hi}}]$ is the 95\% Wilson score interval~\citep{wilson1927probable} on the pooled judgments. Bars are asymmetric and stay within $[0, 100]$ by construction.

\section{Experiments}
We apply the three interventions to a controlled set of image editing tasks designed to isolate specific reference properties.

\subsection{Image Editing Tasks}
\label{subsec:tasks}

% \textbf{Datasets.} 
We evaluate on 2{,}875 editing tasks across five families: object addition (789), object removal (726), human customization (140), color transfer (320), and style transfer (900, split into 450 image pairs each with one realistic and one fictional style). Across all tasks, we write the instruction so that it never names the property we want to study (e.g. color, style, human identity, scene setting) such that this information is only present in the reference image. If that property appears in the generated output image, it must have come from the reference image rather than from the prompt.
% \begin{itemize}
    
    \myparagraph{\textit{Object addition} and \textit{object removal}} reference images are drawn from SUN397 \citep{xiao2010sun}, a scene-recognition dataset; we use two images per category over all 397 indoor and outdoor scene categories, and generate one short noun phrase per image with \texttt{Claude-Opus-4.7} to template into ``add a \emph{<noun>}'' / ``remove the \emph{<noun>}'' instructions (Appendix~\ref{subsec:vlm-sun397}). For removal tasks, we omit images that don't contain a clear object to remove.
    
    \myparagraph{\textit{Human customization}} uses 10 real-human subject images from DreamBench++~\citep{peng2025dreambench++} and edits them to place the humans in different contexts. Each human image is used with $9$ individualized prompts from the dataset and $5$ VLM-generated instructions that are shared across all subjects (Appendix~\ref{subsec:vlm-dreambench-humans}).
    
    \myparagraph{\textit{Color transfer}}  images are one of $8$ solid colors, and the instructions  are ``draw a \textit{<object>} in this color'' for each of $8$ different objects. We use $5$ distinct noise seeds per color-object combination. 
    
    \myparagraph{\textit{Style transfer}} is built from 18 pairs of manually chosen fictional human or animal subjects drawn in different styles, alongside a real human or animal analog image.  The images are  open license images, sourced from Openverse, Openclipart, and iNaturalist (Appendix \ref{sec:compute_and_licenses}). Each pair is combined with $5$ VLM-generated instructions and $5$ noise seeds (Appendix~\ref{subsec:style_transfer_instructions}).
% \end{itemize}

Together, these five families cover a broad slice of the editing behavior FLUX.2 was trained for.

\subsection{What information do the text tokens carry?}
\label{subsec:t2i}

% \textit{T2I Lens probes the intermediate-layer text-token activations to read out what the text tokens have absorbed during a reference-conditioned forward pass.}

We apply T2I Lens to all five task families from Section~\ref{subsec:tasks}. For color transfer, style transfer, and human customization, the judge straightforwardly looks for the reference color, the reference style, or the reference person's identity in the T2I Lens output. For object addition and removal, we ask the judge to look for any distinctive visual aspect of the reference image. Selected qualitative examples are in Figure~\ref{fig:T2I_lens_qual}, with more in Appendix~\ref{subsec:more_T2I_lens}, and quantitative results are in Table~\ref{tab:t2i_lens}.

\begin{figure}[t]
    \centering
    \includegraphics[width=1.0\linewidth]{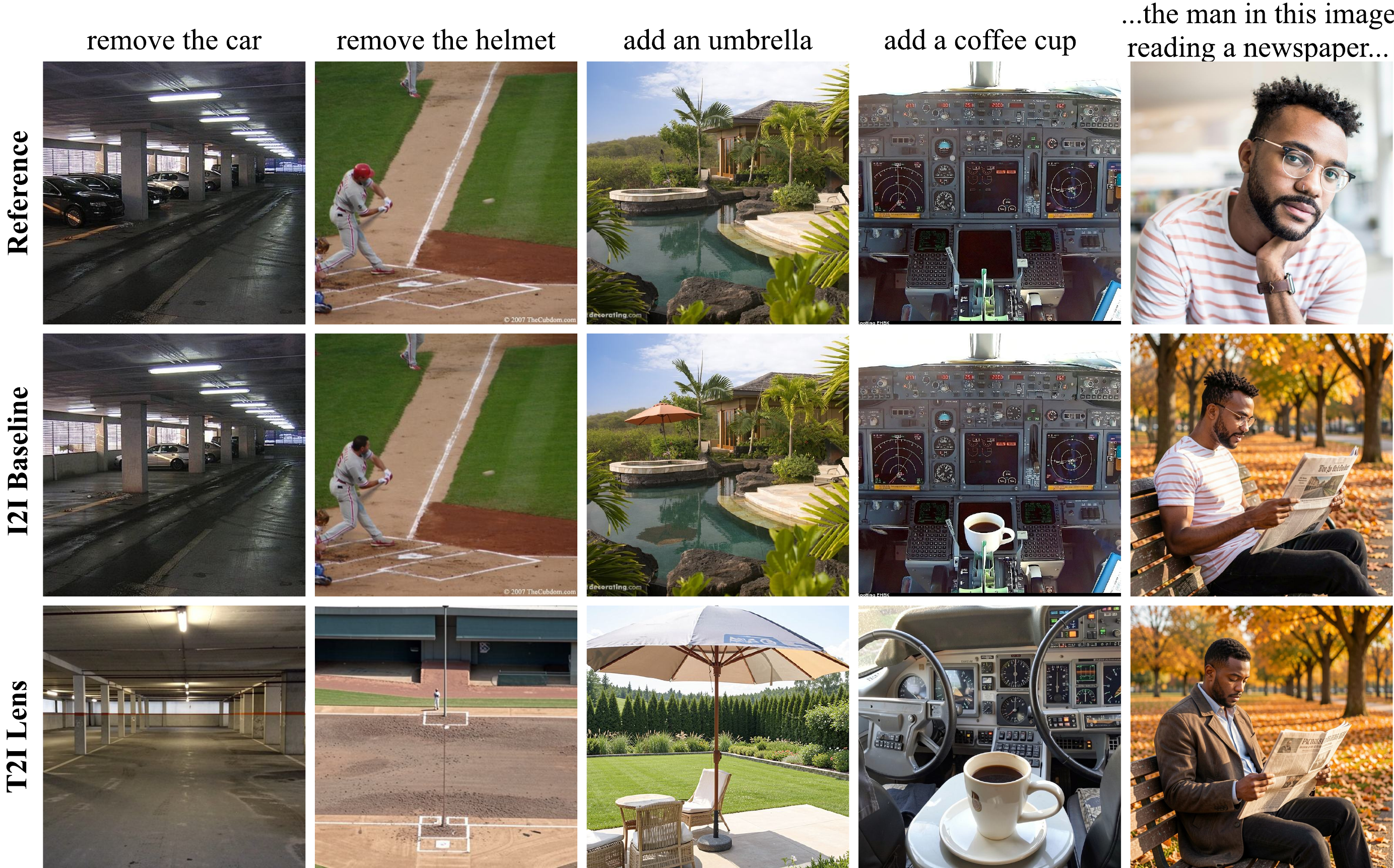}
    \caption{\textbf{T2I Lens}. \textit{I2I Baseline} is generated using the edit instruction on the \textit{Reference} image. We then apply \textit{T2I Lens} to the intermediate-layer \textit{text} tokens of this baseline to visually reveal their encoded information. Despite the  edit instructions  not revealing information about the reference image, the T2I Lens  outputs  generically   match the reference setting (parking garage, sports field, fancy backyard, cockpit), and they contextualize the edit (the umbrella or coffee cup) in that setting. In the rightmost column, exact identity from the reference image such as the human's  hairstyle and clothes are not captured in the text tokens, although general traits like race are still carried over.
    \vspace{-5mm}
    }
    \label{fig:T2I_lens_qual}
\end{figure}

\begin{table}[t]
\centering
\caption{\textbf{T2I Lens Success Rates}. We use a VLM-as-judge to grade each T2I Lens output for the presence or absence of desired properties of the original reference image. Reference properties surface more reliably the simpler and higher-level they are: color, style, and general scene details appear often, while exact human identity is never carried by the text tokens.}
\vspace{3mm}
\label{tab:t2i_lens}
% AUTO-TABLE: t2i_lens START — generated by scripts/build_judge_tables.py
\resizebox{.8\linewidth}{!}{%
% \begin{tabularx}{\linewidth}{l >{\centering\arraybackslash}X >{\centering\arraybackslash}X >{\centering\arraybackslash}X >{\centering\arraybackslash}X >{\centering\arraybackslash}X}
\begin{tabular}{lccccc}
\toprule
 & Color & Style & Object & Object & Human \\
 & Transfer & Transfer & Addition & Removal & Customization \\
\midrule
\makecell[l]{VLM Judge\\Observation Rate (\%)} & $100.0_{-1.2}^{+0.0}$ & $91.8_{-2.9}^{+2.2}$ & $76.8_{-3.1}^{+2.8}$ & $69.0_{-3.5}^{+3.3}$ & $0.0_{-0.0}^{+4.1}$ \\
\bottomrule
% \end{tabularx}
\end{tabular}
}
\vspace{-3mm}
% AUTO-TABLE: t2i_lens END
\end{table}

The text tokens reliably encode the general properties of the reference image (color, style, scene setting), and they reliably fail to encode the exact identity of specific places and people. The fact that the lens recovers anything coherent at all is worth pausing on. The activations we copy come from the middle of a multimodal forward pass, where they have been mixing freely with reference-image tokens through attention, and yet they still serve as valid text input for a fresh generation. Nothing in the FLUX training objective explicitly forces them to stay text-like. The simplest reading is that whatever the text tokens have absorbed has stayed inside the same kind of representation that valid text inputs occupy, and that is also why the bound content is limited to what language can describe.

On their own, T2I Lens results show that the text tokens carry reference content, but like a linear probe, they do not show whether the model is actually using that content when it produces an output, which is the goal of the experiments of the next section.

\subsection{Does the model actually use this binding?}

% \textit{Attention Knockout and I2I-to-I2I Patching causally test whether the reference content present in the text tokens actually drives the generated image.}

\subsubsection{Attention Knockout: When we cut a pathway, which properties break? }
\label{subsec:attention_knockout}

We apply attention knockout to color transfer, style transfer, and human customization. For each result, we grade whether the reference's color, style, or human identity still shows up in the generated image. Quantitative results are in Table~\ref{tab:attention_knockout}, with selected examples (including object addition, which lacks a clean axis to grade) in Figure~\ref{fig:attention_knockout} and more in Appendix~\ref{subsec:more_attention_knockout}.

For color and style, blocking text from attending to the reference is far more damaging than blocking the image from attending to it. The reference content for these properties is first written into the text tokens, which then carry it into the generated image, and the image tokens do not pick these properties up directly from the reference. The pattern flips for human identity, and qualitatively for scene identity as well: blocking the image-side pathway breaks identity preservation, while blocking the text-side pathway barely changes it. Identity, therefore, travels directly from reference to image, skipping the text tokens entirely.

 To confirm that color and style route from the reference through the text tokens, up to a cutoff layer we apply $\text{KO}_{\text{ref}\rightarrow\text{image}}$, and past the cutoff we block all attention into and out of the reference (Appendix~\ref{sec:early_ref_drop}). 
 % $\text{KO}_{\text{ref}\rightarrow\text{image}}$ before the cutoff restricts the reference tokens to only be able to write to the text tokens and not the output image tokens; dropping the reference after the cutoff ensures that the only source of reference information for the output image is what's already encoded in the text tokens.
 With the cutoff set just past the point where the text tokens absorb the reference content, the reference's color and style still appear in the generated image (Figures~\ref{fig:early_ref_drop},\ref{fig:split_grid_color},\ref{fig:split_grid_style}). This is strong evidence for our proposed pathway for reference information, because the knockout leaves the text tokens as the only path the reference could have used. This locality also suggests an efficiency opportunity: once the text tokens hold the reference content, most reference attention can be skipped.

\begin{figure}[t]
    \centering
    \includegraphics[width=1.0\linewidth]{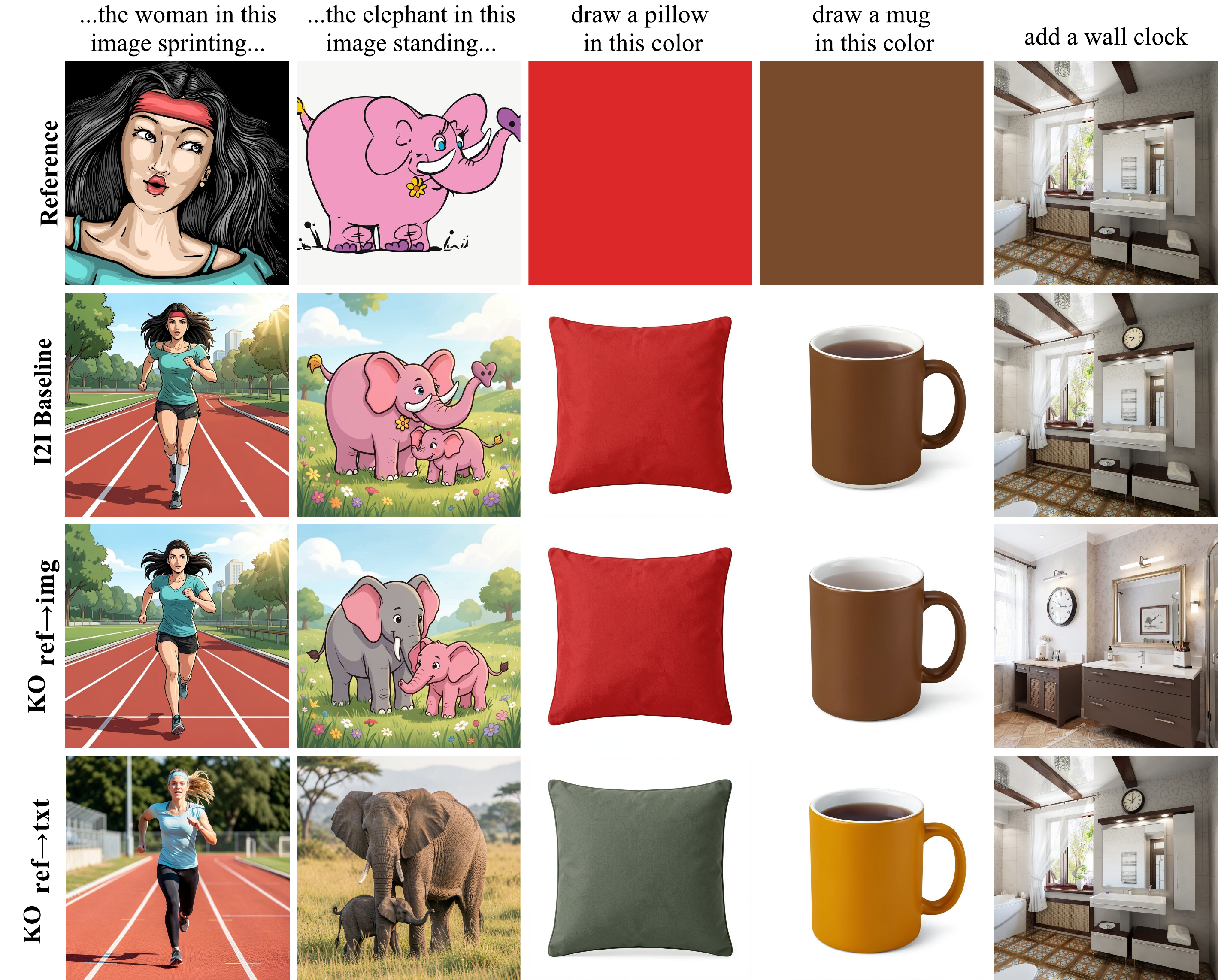}
    \caption{\textbf{Attention Knockout}. \textit{I2I Baseline} is generated using the edit prompt on the \textit{Reference}. In $KO_{ref\rightarrow img}$, we knock out attention paid by the image tokens to the reference tokens and repeat the editing task, and similarly for $KO_{ref\rightarrow txt}$. In the first four columns, $KO_{ref\rightarrow img}$ does not prevent the reference's color or style from appearing in the generated output, while $KO_{ref\rightarrow txt}$ fully prevents it, keeping the pillow gray instead of red and using a real elephant instead of the clipart-style one in the reference. The opposite effect occurs for preserving exact scene identity: in the last column, $KO_{ref\rightarrow txt}$ leaves the bathroom unchanged, while $KO_{ref\rightarrow img}$ completely alters its identity.
    \vspace{-3mm}}
    \label{fig:attention_knockout}
\end{figure}

\subsubsection{I2I-to-I2I Patching: When we copy text tokens, what properties travel with them?}

The setup is the one we described earlier: a source I2I edit and a target I2I edit that share the same instruction but use different reference images, with text-token activations copied from source to target. If the text tokens are causally carrying the source reference's color, style, or identity, the target output should pick that property up. We construct $448$ such patch pairs for color transfer ($8 \times 7$ source-target color pairs across $8$ instructions), $450$ for style transfer ($18$ subjects $\times$ $5$ instructions $\times$ $5$ seed pairs), and $450$ for human customization ($10 \times 9$ source-target subject pairings across $5$ shared instructions). Source and target always use different noise seeds, so shared noise cannot drive the result.

Quantitative results are in Table~\ref{tab:i2i_to_i2i}, with selected examples in Figure~\ref{fig:i2i_to_i2i} and more in Appendix~\ref{subsec:more_i2i_to_i2i_patching}. The text tokens reliably carry color and style from the source reference to the target output, but carry nothing about identity. In many cases the source reference's color or style is the only attribute of the target output that visibly changes, suggesting the text tokens play a narrow, targeted role in routing reference content rather than acting as a general-purpose carrier.

\begin{figure}[t]
    \centering
    \includegraphics[width=1.0\linewidth]{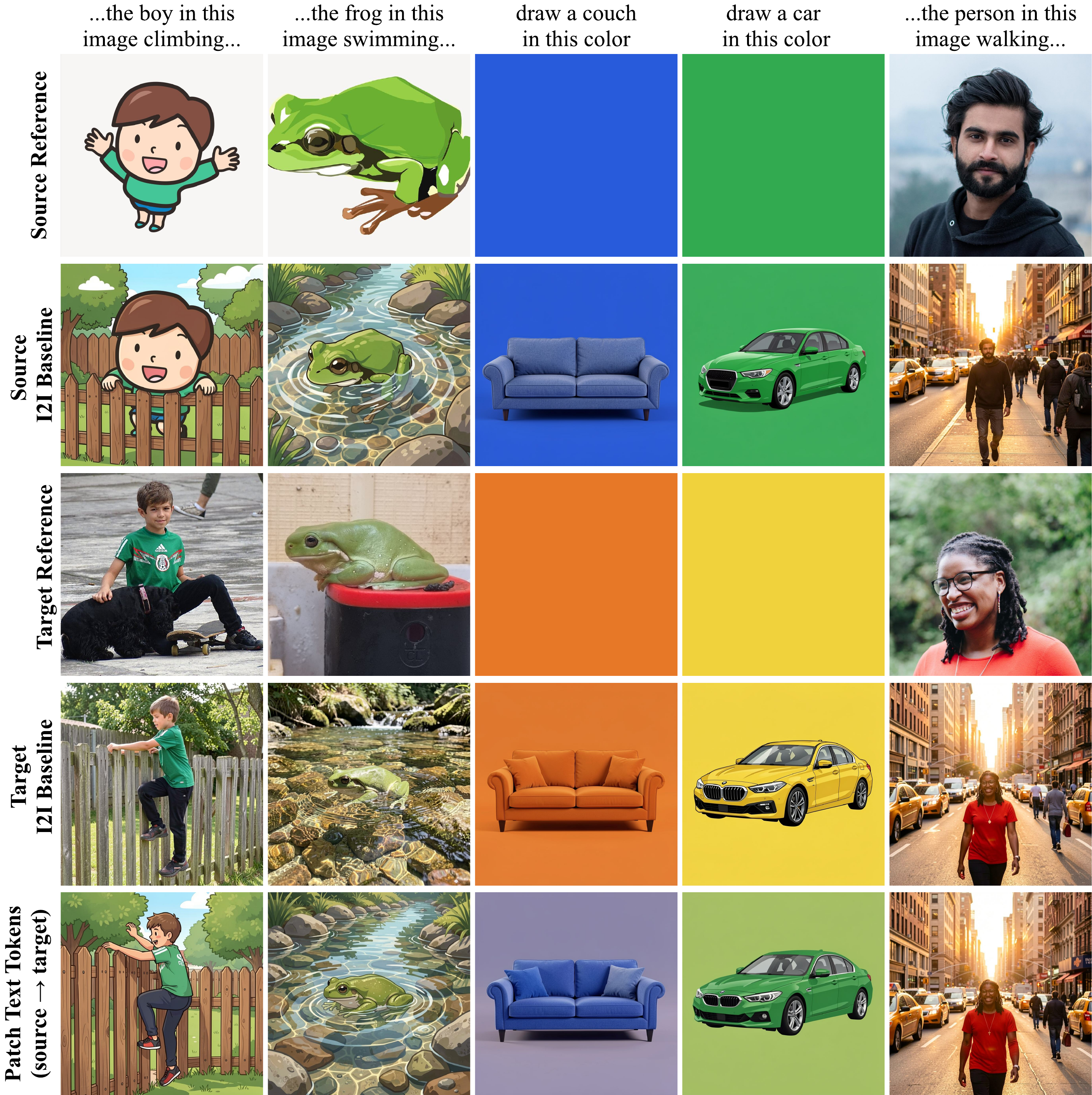}
    \caption{\textbf{I2I-to-I2I Patching}. The \textit{Source I2I Baseline} is generated using the edit on the \textit{Source Reference}, and similarly for the \textit{Target I2I Baseline}. Text-token activations are patched from the Source to Target I2I generations in corresponding layers, yielding the \textit{Target I2I Patched} output. Patching successfully transfers the style of the reference onto the image of the boy in the first column, while preserving other details like the stripes on his shirt and the shape of the fence. Patching also brings the reference blue and reference green onto couch and car, respectively. The identity of the woman in the last column, in contrast, is unaffected by patching. Patching text tokens causally transfers color and style from source reference to target, but not human identity.}
    \label{fig:i2i_to_i2i}
\end{figure}

\begin{table}[t]
\centering
\caption{\textbf{Attention Knockout Success Rates}. For each knockout setting, we report the percentage of tasks where the reference property (color, style, or identity) is destroyed. $KO_{\text{ref}\rightarrow\text{text}}$ destroys color and style transfer, while $KO_{\text{ref}\rightarrow\text{image}}$ leaves them mostly intact; the pattern flips for human identity.}
\label{tab:attention_knockout}
% AUTO-TABLE: attention_knockout START — generated by scripts/build_judge_tables.py
\vspace{3mm}
\resizebox{.85\linewidth}{!}{%
% \begin{tabularx}{\linewidth}{l >{\centering\arraybackslash}X >{\centering\arraybackslash}X >{\centering\arraybackslash}X >{\centering\arraybackslash}X >{\centering\arraybackslash}X}
\begin{tabular}{lccccc}
% \begin{tabularx}{\linewidth}{l >{\centering\arraybackslash}X >{\centering\arraybackslash}X >{\centering\arraybackslash}X}
\toprule
 & Color Transfer (\%) & Style Transfer (\%) & Human Customization (\%) \\
\midrule
KO\textsubscript{ref$\rightarrow$text} & $86.2_{-4.2}^{+3.3}$ & $97.3_{-1.9}^{+1.1}$ & $45.6_{-9.9}^{+10.3}$ \\
KO\textsubscript{ref$\rightarrow$image} & $13.4_{-3.3}^{+4.2}$ & $16.4_{-3.1}^{+3.7}$ & $76.7_{-9.7}^{+7.5}$ \\
\bottomrule
% \end{tabularx}
\end{tabular}
}
% AUTO-TABLE: attention_knockout END
\end{table}

\subsection{Where in the text sequence does the binding live?}
\label{subsec:padding}

\begin{figure}[t]
    \centering
    \includegraphics[width=1.0\linewidth]{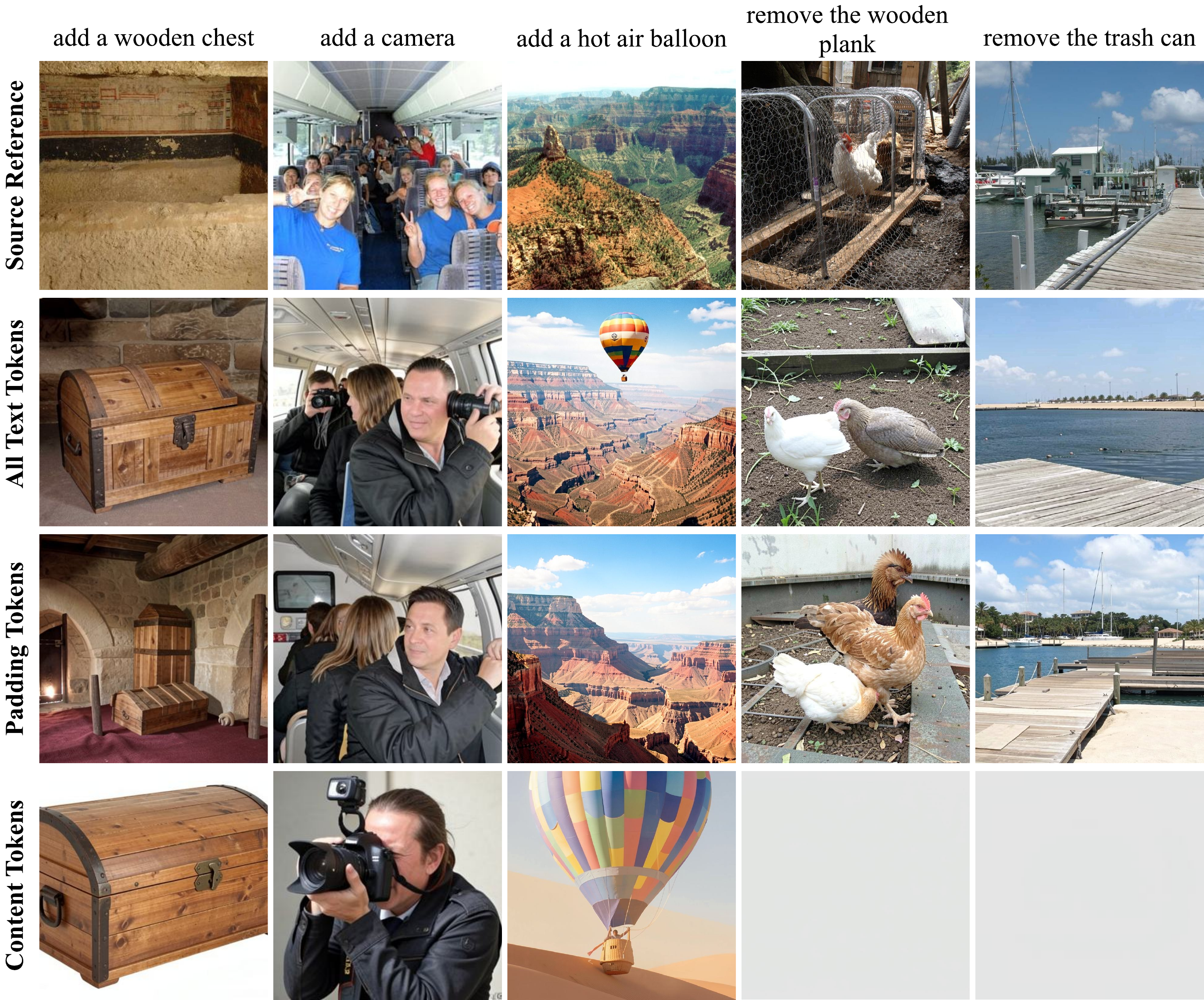}
    \caption{\textbf{T2I Lens on only padding or content text tokens.} We repeat our T2I Lens technique on all tasks, but only patching over a subset of the text tokens. The padding tokens on their own consistently encode reference scene information. Unlike T2I Lens on all text tokens, T2I Lens on the padding tokens does not always include the edit: there is no camera or hot air balloon in the second and third examples. In contrast, the content tokens focus solely on the edit instruction, not including any reference details, which results in empty images when the instruction is to remove an object. 
    \vspace{-3mm}
    }

    \label{fig:t2i_lens_padding}
\end{figure}

The text sequence has two types of tokens: the content tokens, which encode the edit instruction, and the padding tokens to fit the standard sequence length. To find out which of these holds the bound reference content, we run two probes. First, we use T2I Lens on only the padding tokens or only the content tokens, patching certain token indices but not others (Figure~\ref{fig:t2i_lens_padding}). The padding tokens on their own consistently encode reference scene information, while the content tokens focus solely on the edit instruction. Second, we run two more variants of I2I-to-I2I patching: one that patches only the padding tokens, and one that patches only the content tokens (Table~\ref{tab:i2i_to_i2i}, with examples in Figure~\ref{fig:i2i_patching_padding}). Patching just the padding transfers color and style at nearly the same rate as patching all text tokens; patching just the content transfers almost nothing. The reference content is therefore held on the padding tokens, not stacked on top of the content tokens that already describe the edit. The same padding-only or content-only split applies to attention knockout (Appendix~\ref{sec:attention-knockout-full}).

\begin{table}[t]
\centering
\caption{\textbf{I2I-to-I2I Patching Results}. Patching text tokens reliably transfers the source reference's color or style to the target, with only a slight drop in success rate when patching only the text padding tokens. The success rate drops to almost zero when only the text content tokens (the actual edit instruction) are patched. Patching text has no effect on human identity.}
\label{tab:i2i_to_i2i}
\vspace{3mm}
\resizebox{.95\linewidth}{!}{%
% \begin{tabularx}{\linewidth}{l >{\centering\arraybackslash}X >{\centering\arraybackslash}X >{\centering\arraybackslash}X >{\centering\arraybackslash}X >{\centering\arraybackslash}X}
\begin{tabular}{lccc}
% AUTO-TABLE: i2i_to_i2i START — generated by scripts/build_judge_tables.py
% \begin{tabularx}{\linewidth}{l >{\centering\arraybackslash}X >{\centering\arraybackslash}X >{\centering\arraybackslash}X}
\toprule
 & Color Transfer (\%) & Style Transfer (\%) & Human Customization (\%) \\
\midrule
Text Tokens (All) & $92.0_{-2.9}^{+2.2}$ & $87.6_{-3.4}^{+2.7}$ & $0.7_{-0.4}^{+1.3}$ \\
Text Tokens (Padding Only) & $85.0_{-3.6}^{+3.0}$ & $82.0_{-3.8}^{+3.3}$ & $0.0_{-0.0}^{+0.8}$ \\
Text Tokens (Content Only) & $0.4_{-0.3}^{+1.2}$ & $2.9_{-1.2}^{+2.0}$ & $0.4_{-0.3}^{+1.2}$ \\
\bottomrule
% \end{tabularx}
\end{tabular}
}
% AUTO-TABLE: i2i_to_i2i END
\end{table}

\subsection{Limitations}
\label{subsec:limitations}

The role we describe varies with the task. Color and style are two properties for which we have shown that the text tokens are causally responsible, but we have not exhaustively cataloged every behavior this mechanism participates in. Even for color and style, the attention-knockout and I2I-to-I2I patching results are not perfect, which suggests the model has secondary or backup pathways for transferring these properties. Mapping those out is beyond the scope of this paper, which is focused on establishing the role of the text tokens rather than on reverse-engineering the full mechanism.

Our analysis is also restricted to FLUX. Among general open editing models, only Qwen-Image-Edit shares the architectural choice of treating reference-image tokens the same way as output-image tokens, so that both attend to the text and to each other. But as we discuss in Section~\ref{subsec:image_editing}, the Qwen2.5-VL encoder that produces Qwen-Image-Edit's text embeddings already takes the reference image as one of its inputs, so reference content is in the text stream from the very start by construction. The routing question we ask of FLUX is therefore not a meaningful question to ask of Qwen-Image-Edit.

\section{Conclusion}
We study how a single set of FLUX.2 weights routes information between text tokens and reference-image tokens during in-context editing. Using three causal interventions on the multimodal attention stream, T2I Lens, Attention Knockout, and I2I-to-I2I Patching, we find that an implicit vision-language binding emerges between the text tokens and the reference image. Reference properties such as the color and style of a scene are first written into the text padding tokens, which then carry them into the generated image. This makes the padding length act as an implicit capacity hyperparameter for cross-modal binding. Pixel-exact properties like identity bypass the text tokens and travel directly through image-to-image attention.

% The split is principled, and it mirrors the architectural choices of modern in-context image generation models that separate semantic and appearance pathways by design. 
This split of information is principled, and it mirrors the modern explicit two-encoder design of architectures like Qwen-Image-Edit, which routes the reference image through a vision-language encoder for semantics and a VAE for appearance. We hope our findings spark novel, more efficient, and interpretable generative architectures, rather than treating the unified attention stream as a black box that conditioning enters and an image leaves.
 
\begin{ack}
This research was supported by ARL grant \#W911NF-24-2-0069, MIT-IBM Watson AI Lab grant \#W1771646, and Hyundai Motor Company. RG is funded by NSF grant \#2403303. 
\end{ack}

\medskip
{
\small
\bibliographystyle{plainnat}
\bibliography{references}
}

%%%%%%%%%%%%%%%%%%%%%%%%%%%%%%%%%%%%%%%%%%%%%%%%%%%%%%%%%%%%
\newpage
\appendix
\renewcommand{\thefigure}{\thesection\arabic{figure}}
\renewcommand{\thetable}{\thesection\arabic{table}}
\counterwithin*{figure}{section}
\counterwithin*{table}{section}
\Large{\textbf{Appendix}}
\normalsize

\section{Specific choice of layers in FLUX.2 Klein 9B to patch}
\label{sec:mm7_single9}

For the experiments in the main paper, whenever we patch text activations from an I2I generation, it is always from after the $8$th double stream block, or the $10$th single stream block if it is a color transfer task. In this section, we justify this choice by showing a grid of results for various tasks on T2I Lens, demonstrating what would happen if we patched activations from a different layer. 

\subsection{T2I Lens for Scenes, Style, and Humans}

\begin{figure}[h]
    \centering
    \includegraphics[width=1\linewidth]{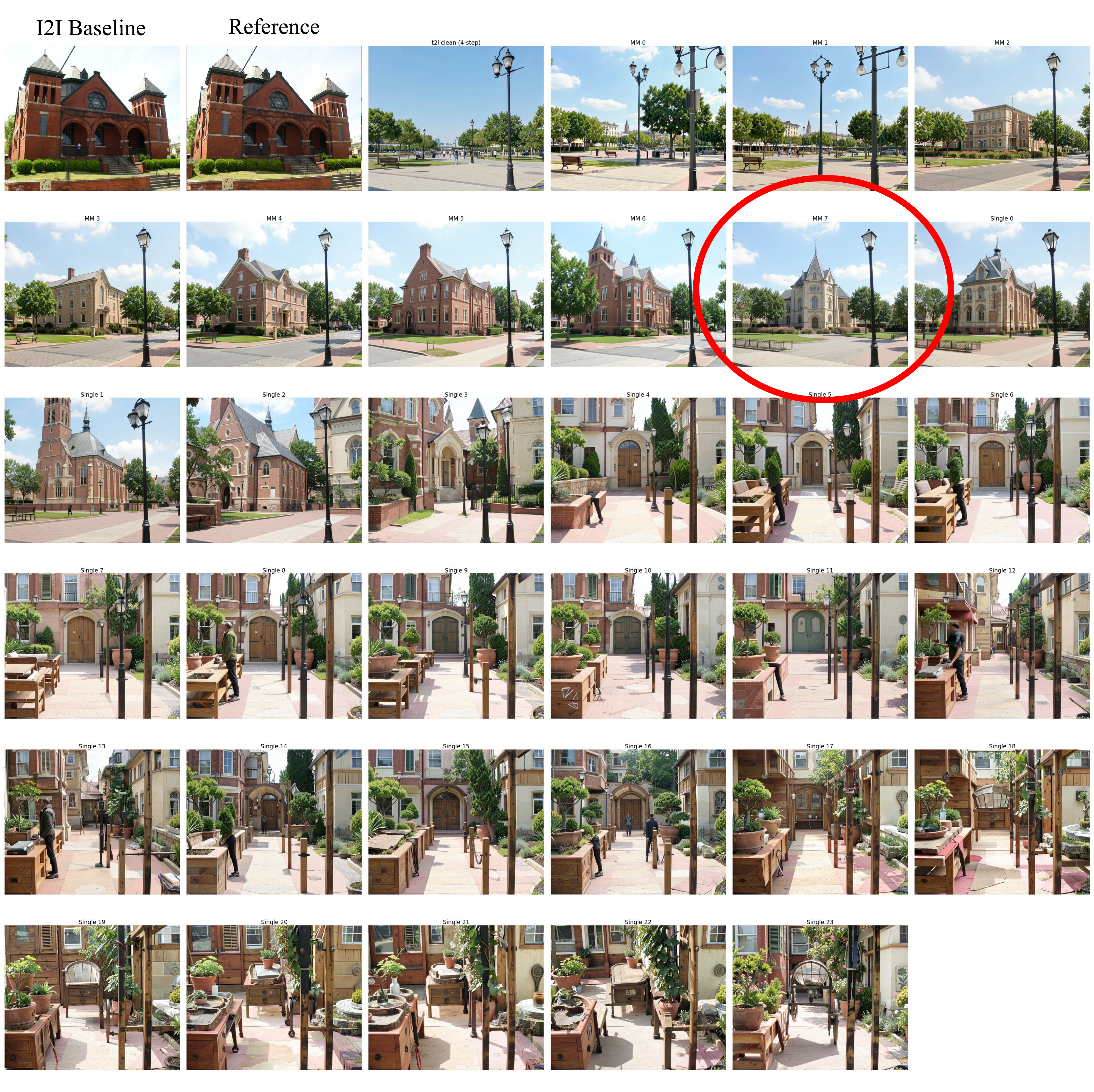}
    \caption{\textbf{T2I Lens patching text from different layers for adding a lamp post to a church.} The lamp post is fully contextualized to be in a setting with a church by the circled 8th double stream block. In later layers, the church in the background transforms into a less coherent building structure.  }
    \label{fig:scenes_grid1}
\end{figure}

\begin{figure}[h]
    \centering
    \includegraphics[width=1\linewidth]{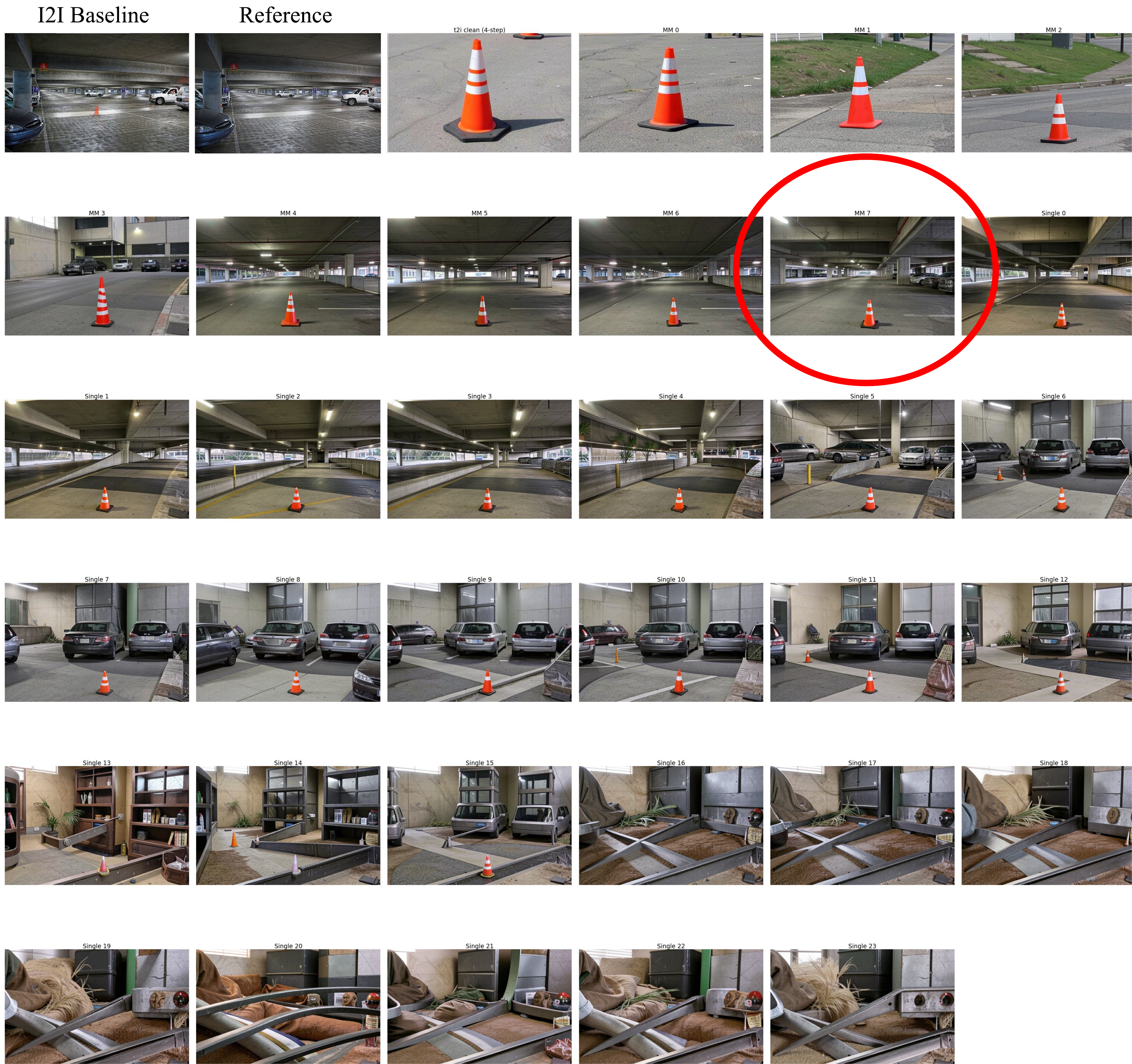}
    \caption{\textbf{T2I Lens patching text from different layers for adding a traffic cone to a parking garage.} Both the parking garage and the traffic cone are fully instantiated in the Lens output image by the circled 8th double stream block. In later layers the background changes to no longer be a parking garage.  }
    \label{fig:scenes_grid2}
\end{figure}

\begin{figure}[h]
    \centering
    \includegraphics[width=1\linewidth]{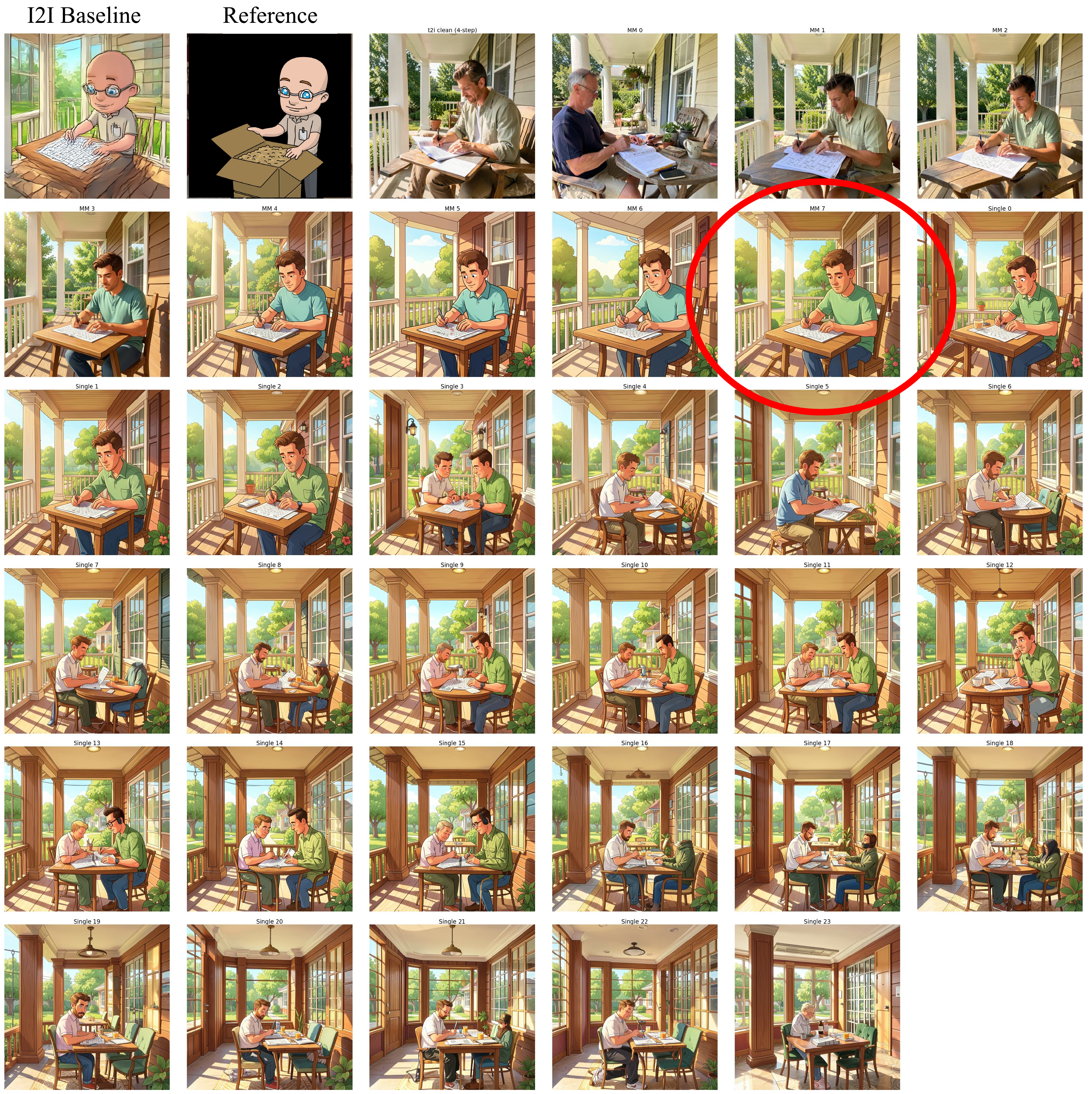}
    \caption{\textbf{T2I Lens patching text from different layers for a man solving a crossword}. The man is fully drawn in a fictional style by the circled 8th double stream block. }
    \label{fig:style_grid1}
\end{figure}

\begin{figure}[h]
    \centering
    \includegraphics[width=1\linewidth]{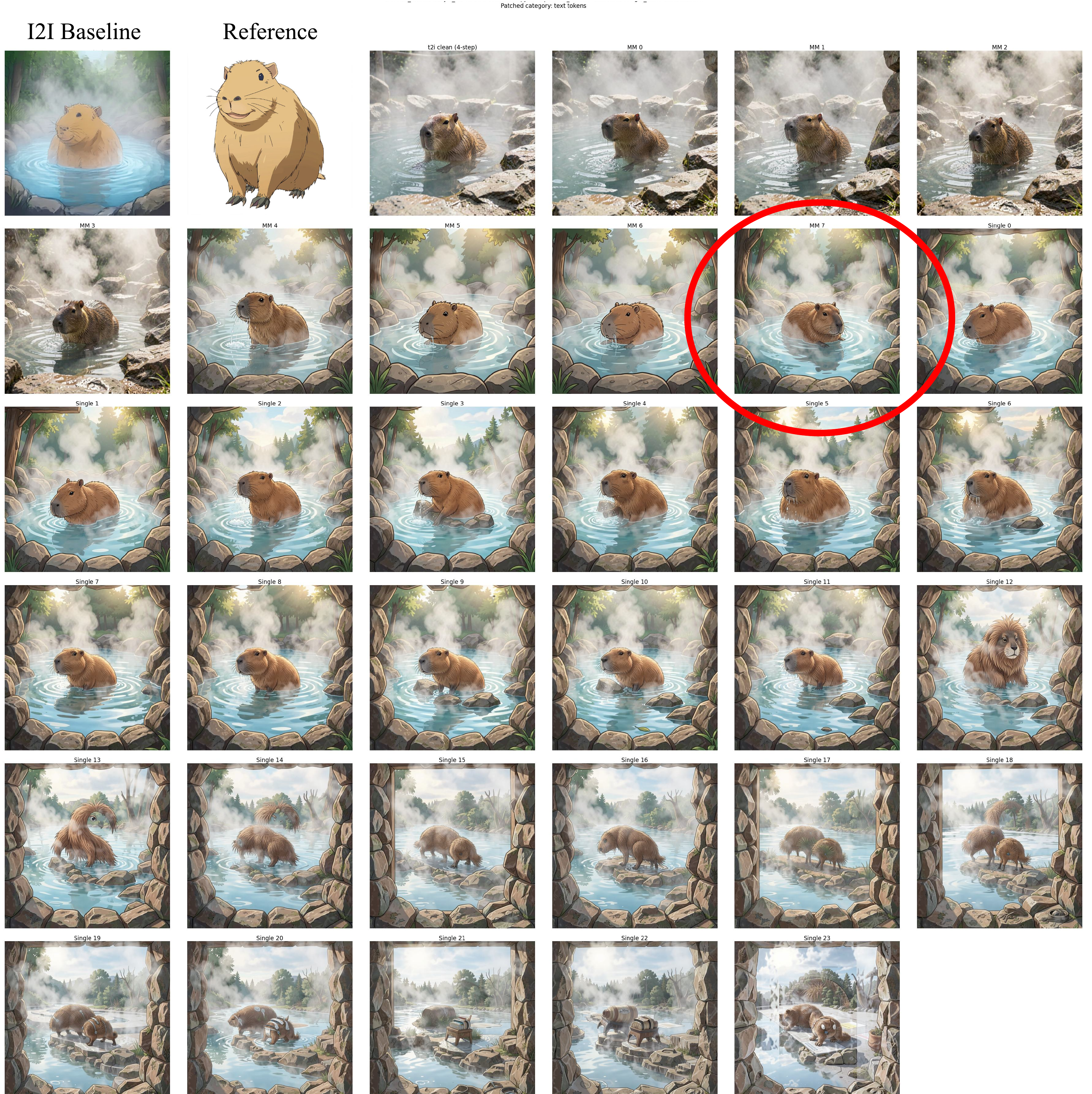}
    \caption{\textbf{T2I Lens patching text from different layers for a capybara swimming in a hot spring}. The capybara is fully drawn in a fictional style by the circled 8th double stream block. In later layers, this fictional style disappears again, and the capybara is also less clear.   }
    \label{fig:style_grid2}
\end{figure}

\begin{figure}[h]
    \centering
    \includegraphics[width=1\linewidth]{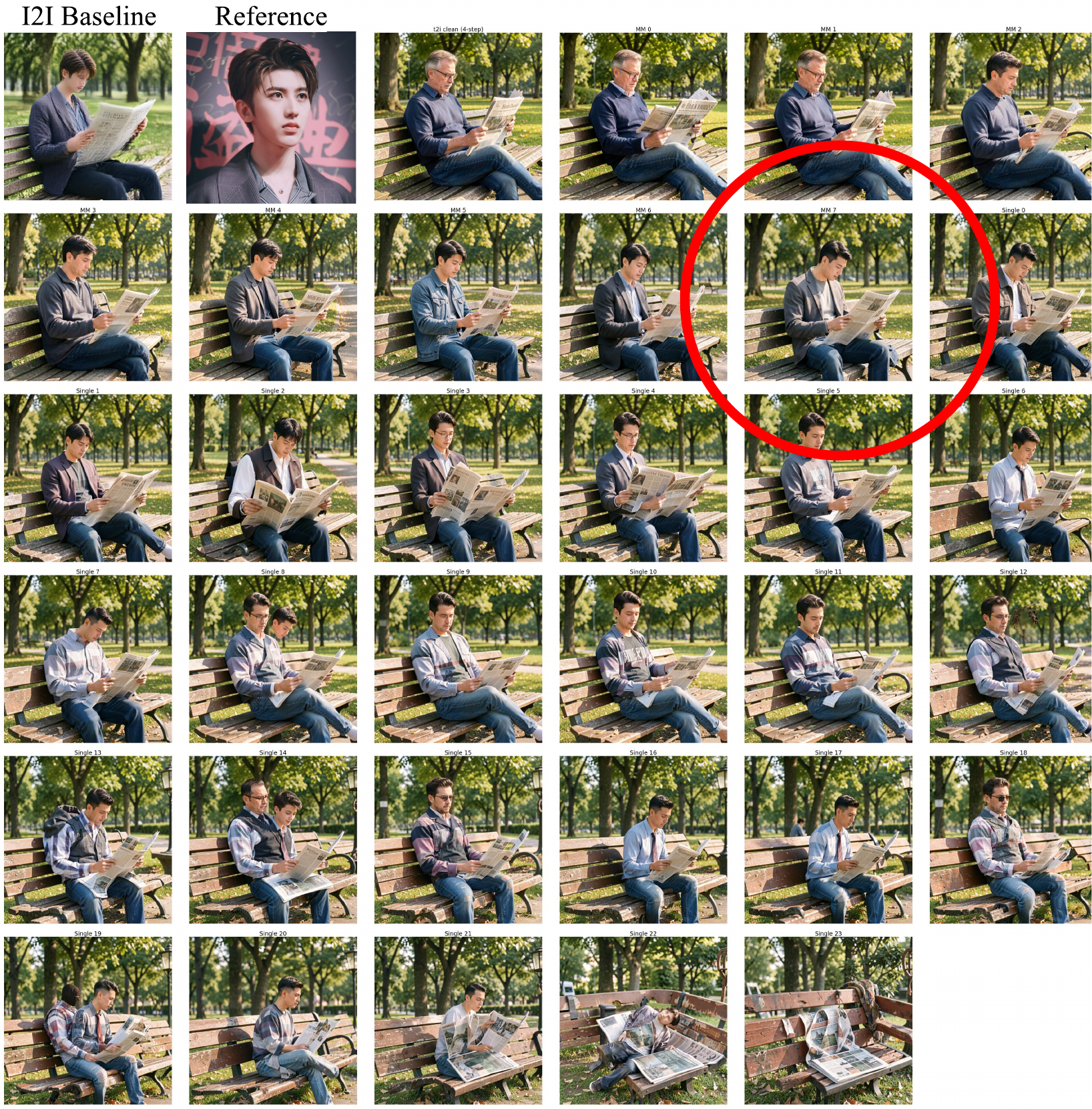}
    \caption{\textbf{T2I Lens patching text from different layers for a man reading a newspaper }. The man is correctly portrayed as Asian by the 8th double stream block. In significantly earlier and later layers, the man is portrayed as white instead.}
    \label{fig:humans_grid}
\end{figure}

Figures \ref{fig:scenes_grid1}, 
\ref{fig:scenes_grid2}, \ref{fig:style_grid1},
\ref{fig:style_grid2}, \ref{fig:humans_grid} show the T2I Lens results of patching different layers on the tasks ``Add a lamp post,'' (to a church)  ``Add a traffic cone,'' (to a parking garage) ``A photograph of the man in this image solving a crossword on a sunny front porch,'' ``A photograph of the capybara in this image bathing in a steamy hot spring,'' and ``A photograph of the man in this image reading a newspaper on a park bench,'' respectively. We see that by around the 6-8th double stream block (the red circled block is the 8th double stream block), key properties of the reference are already manifest in the T2I Lens outputs: the lamp post is in front of a church, the traffic cone is in a parking garage, the man  solving the crossword is styled in a fictional manner, the capybara is styled in a fictional manner, and the man reading the newspaper is Asian. Notice that none of these reference properties are mentioned or implied by the edit instruction. We therefore select the 8th double stream block as the point at which the text tokens should have contextualized the reference. Later layers degrade in semantic clarity.

\clearpage

\clearpage

\clearpage

\clearpage

\clearpage
\clearpage

\subsection{T2I Lens for Color} 
\label{subsec:t2i_color}
\begin{figure}[h]
    \centering
    \includegraphics[width=1\linewidth]{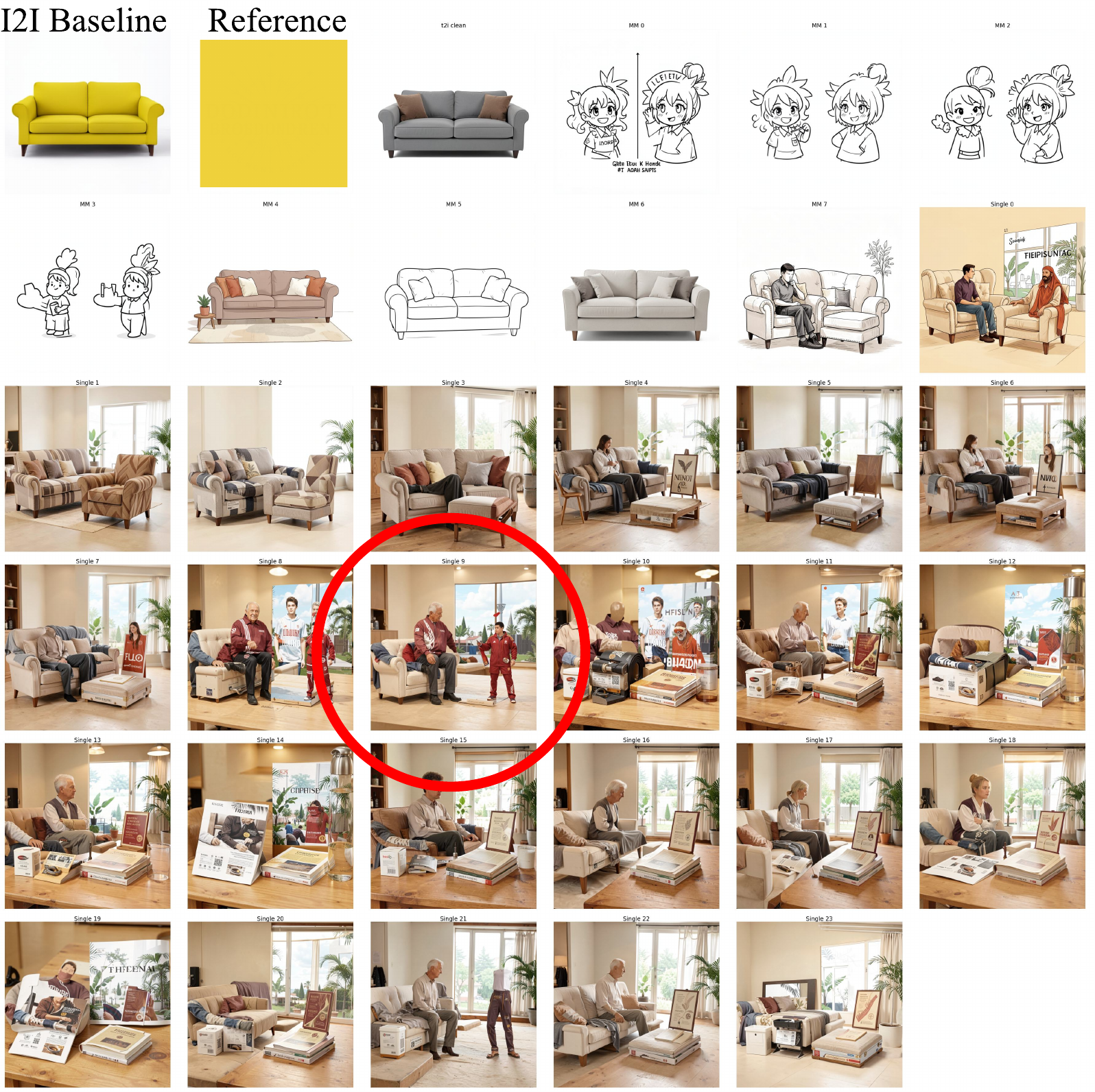}
    \caption{\textbf{Standard T2I Lens on color tasks} results in a very vague background hue of the reference color (in this case, yellow) by the 10th single stream block (circled), but it is difficult to tell from looking at the image for any one patched layer result in isolation. }
    \label{fig:input_four_step_grid}
\end{figure}

\begin{figure}[h]
    \centering
    \includegraphics[width=1\linewidth]{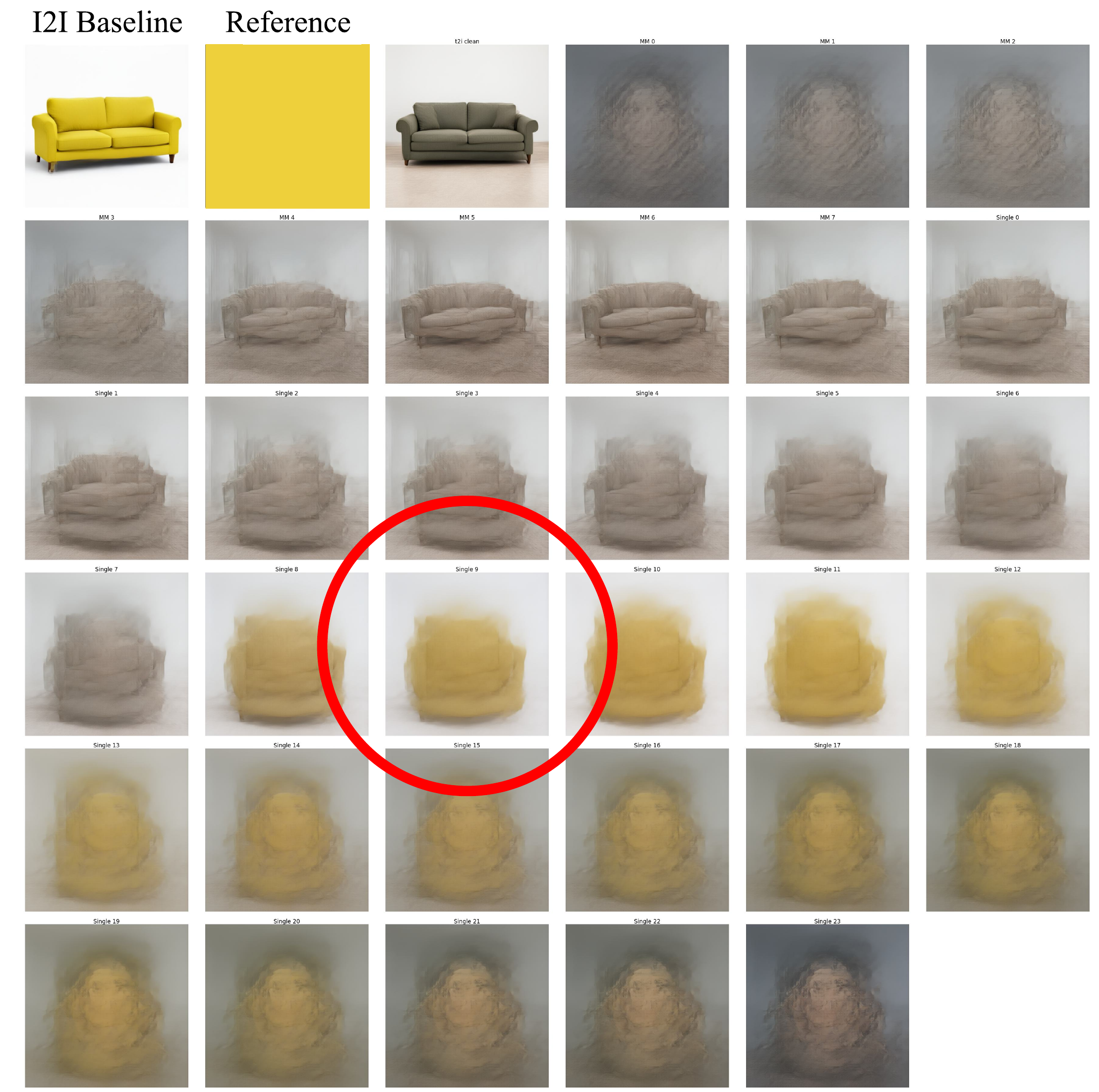}
    \caption{\textbf{Modified T2I Lens on color tasks}. By patching to the same layer and using single-step generation instead, the T2I Lens images drop in quality, but the patched reference color stands out much more; the yellow is clearly present by the circled 10th single stream block. }
    \label{fig:same_layer_single_step}
\end{figure}

For the T2I Lens experiments, we treat color tasks differently from the other tasks in two ways. The first is by selecting a different layer to patch the text tokens from: the 10th single stream block (a much later layer than the 8th double stream block). The second is that we patch the activations to the same layer in the unconditional T2I and use one-step generation, rather than patching them to the input text embeddings and using four-step generation. The difference is unimportant in terms of the interpretation: either way, if the color of the reference appears in the T2I Lens image, then it means that it was encoded in the text tokens at that point in the computation. We find that with input layer, 4-step generation like in Figure \ref{fig:input_four_step_grid}, the reference color still exists in the T2I images as a slight tinge, but without the context of the patching results from other layers, it's difficult to tell whether the color is intentionally added or if it would naturally belong in the image. In contrast, we empirically find that for same-layer, single-step generation T2I Lens as in Figure \ref{fig:same_layer_single_step}, the emergence of the reference color in the T2I Lens is extremely clear and consistent between tasks. 

% Interestingly, even for tasks that do not explicitly involve color transfer but whose reference images involve a strong color tone, like a reference image of the ocean (Figure \ref{fig:ocean_color}), we find a similar emergence of the background color in the T2I Lens images around the same layer, reinforcing that the $10$th single stream block is where some reference color mechanism takes place in the text tokens. We hypothesize that color is transferred through a different mechanism than the other reference properties, since it binds later in the network and at a different block type.
\clearpage

\clearpage

\clearpage

% \begin{figure}[h]
%     \centering
%     \includegraphics[width=1\linewidth]{Figures/color_grid3.pdf}
%     \caption{\textbf{Modified T2I Lens with implicit color transfer}. Even when the prompt (``Add an armchair'') does not explicitly mention color, we still observe a more vibrant blue in the T2I Lens images around the circled 10th single stream block. }
%     \label{fig:ocean_color}
% \end{figure}

\clearpage
\section{Dropping the Reference Image After the Binding Layer}
\label{sec:early_ref_drop}

To test our claim that the reference's color and style reach the output through the text
tokens, we remove the reference image's information itself partway through the model and check whether
the content still transfers. The knockouts in Section~\ref{subsec:attention_knockout}
sever the same attention edge at every layer; here we vary the knockout by layer: up to a
cutoff layer we apply $\text{KO}_{\text{ref}\rightarrow\text{image}}$, and past the cutoff
we block all attention into and out of the reference
($\text{KO}_{\text{ref}\rightarrow\text{image}}$,
$\text{KO}_{\text{ref}\rightarrow\text{text}}$,
$\text{KO}_{\text{image}\rightarrow\text{ref}}$, and
$\text{KO}_{\text{text}\rightarrow\text{ref}}$), removing it from the rest of the
computation. 
These two knockouts isolate the two sub-processes of the binding mechanism; up to the cutoff, we block only $\text{KO}_{\text{ref}\rightarrow\text{image}}$, leaving the reference free to write into the text tokens but preventing it from reaching the image tokens directly. Past the cutoff, we drop the reference entirely, so any content that did not make it into the text tokens by the cutoff layer has no further opportunity to influence the output. If reference attributes survive this double intervention, it means they were written into the text tokens before the cutoff and carried forward from there, with no remaining pathway through the reference image itself.
We place the cutoff at the binding layers from
Appendix~\ref{sec:mm7_single9}: the $8$th double stream block for style and scene tasks,
and the $10$th single stream block for color tasks. 

\begin{figure}[h]
    \centering
    \includegraphics[width=1.0\linewidth]{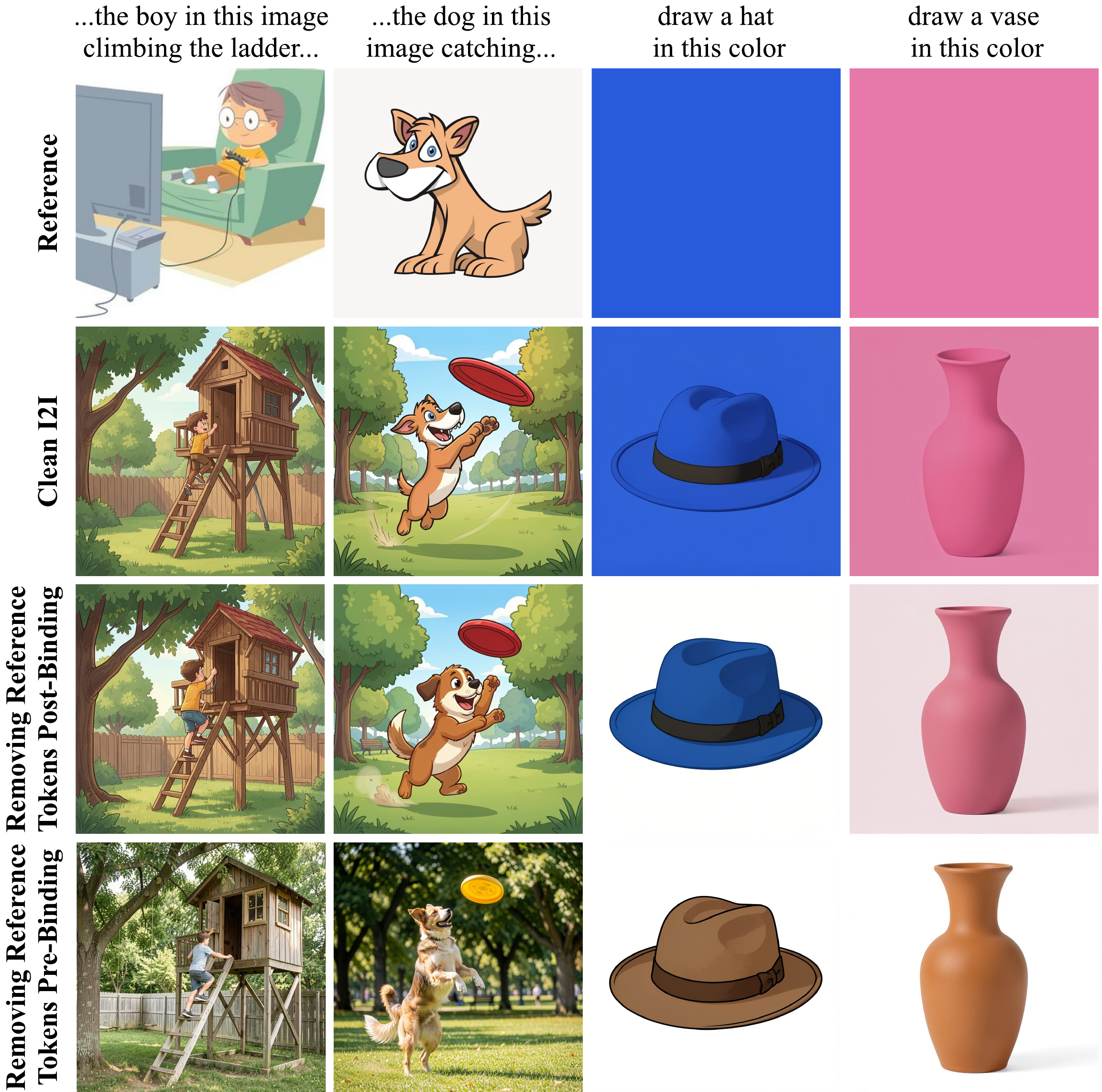}
    \caption{\textbf{Dropping the reference image after the binding layer.} \textit{Clean I2I} is the reference-conditioned edit, and \textit{Removing Reference Tokens Pre-Binding} drops the
    reference entirely from the first layer (it's a T2I). In contrast, \textit{Removing Reference Tokens Post-Binding}  applies
    $KO_{\text{ref}\rightarrow \text{img}}$ in every layer up to some fixed cutoff layer and fully cuts off the reference tokens from the other tokens  past that cutoff layer; the cutoff layer is chosen to be just after where we believe vision-language binding  occurs. Despite this aggressive blocking, \textit{Removing Reference Tokens Post-Binding} still
    reproduces the reference's cartoon style (the boy and dog examples) and color (the blue hat and 
    pink vase examples), matching the \textit{Clean I2I} rather than the reference-free
    \textit{Removing Reference Tokens Pre-Binding}. By the cutoff layer, the text tokens have already absorbed the
    reference content and carry it to the output on their own.}
    \label{fig:early_ref_drop}
\end{figure}

Figure~\ref{fig:early_ref_drop} shows
the result on selected color and style tasks.
Even though the reference is cut off partway through the model, its color and style still
appear in the output, providing strong evidence for our claim that the  reference
writes its color and style into the text tokens early, and the text tokens carry those
properties to the generated image on their own, without the direct reference-to-image
route.

\begin{figure}[h]
    \centering
    \includegraphics[width=1\linewidth]{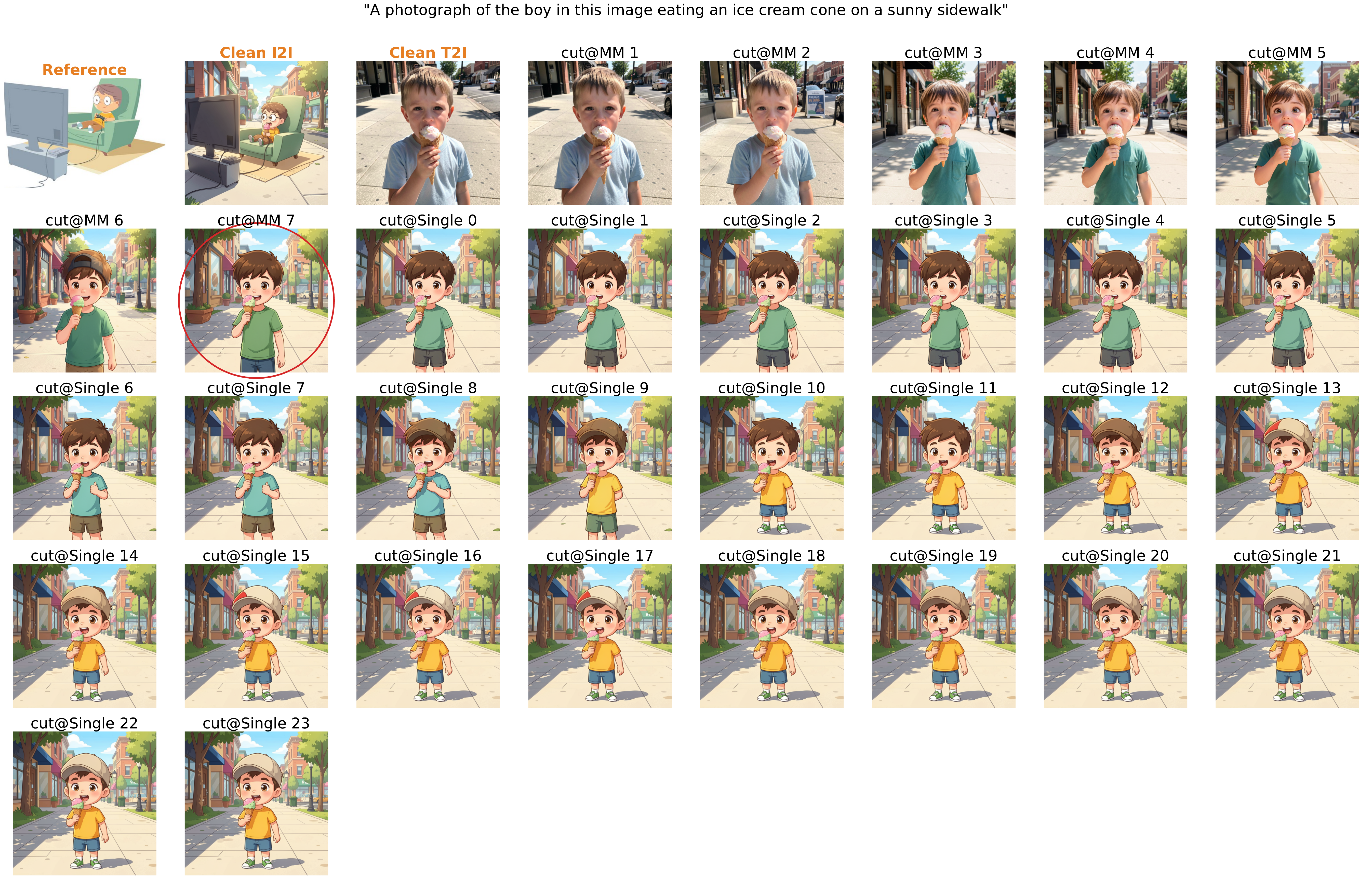}
    \caption{\textbf{Sweeping the reference-drop cutoff on a style transfer task.} Each
    cell runs the same experiment of Figure \ref{fig:early_ref_drop}  on a style edit (a cartoon boy eating ice cream), with
    the cutoff moved to a progressively later block. The cartoon style is absent when the
    cutoff is early and is fully present once the cutoff passes the 8th double stream
    block, matching the style binding layer in Appendix~\ref{sec:mm7_single9}.}
    \label{fig:split_grid_style}
\end{figure}

Sweeping the reference-drop cutoff across all $32$ blocks measures where the reference
writes its content into the text tokens, independently of the T2I Lens used to locate the
binding layers in Appendix~\ref{sec:mm7_single9}.
In Figures~\ref{fig:split_grid_color} and~\ref{fig:split_grid_style}, a cutoff that is too
early removes the reference before its content has reached the text tokens, so the color
or style is lost; once the cutoff is late enough, the property transfers in full. The
transition happens at those same binding layers, confirming the layer choice.

\clearpage

\clearpage
\begin{figure}[h]
    \centering
    \includegraphics[width=1\linewidth]{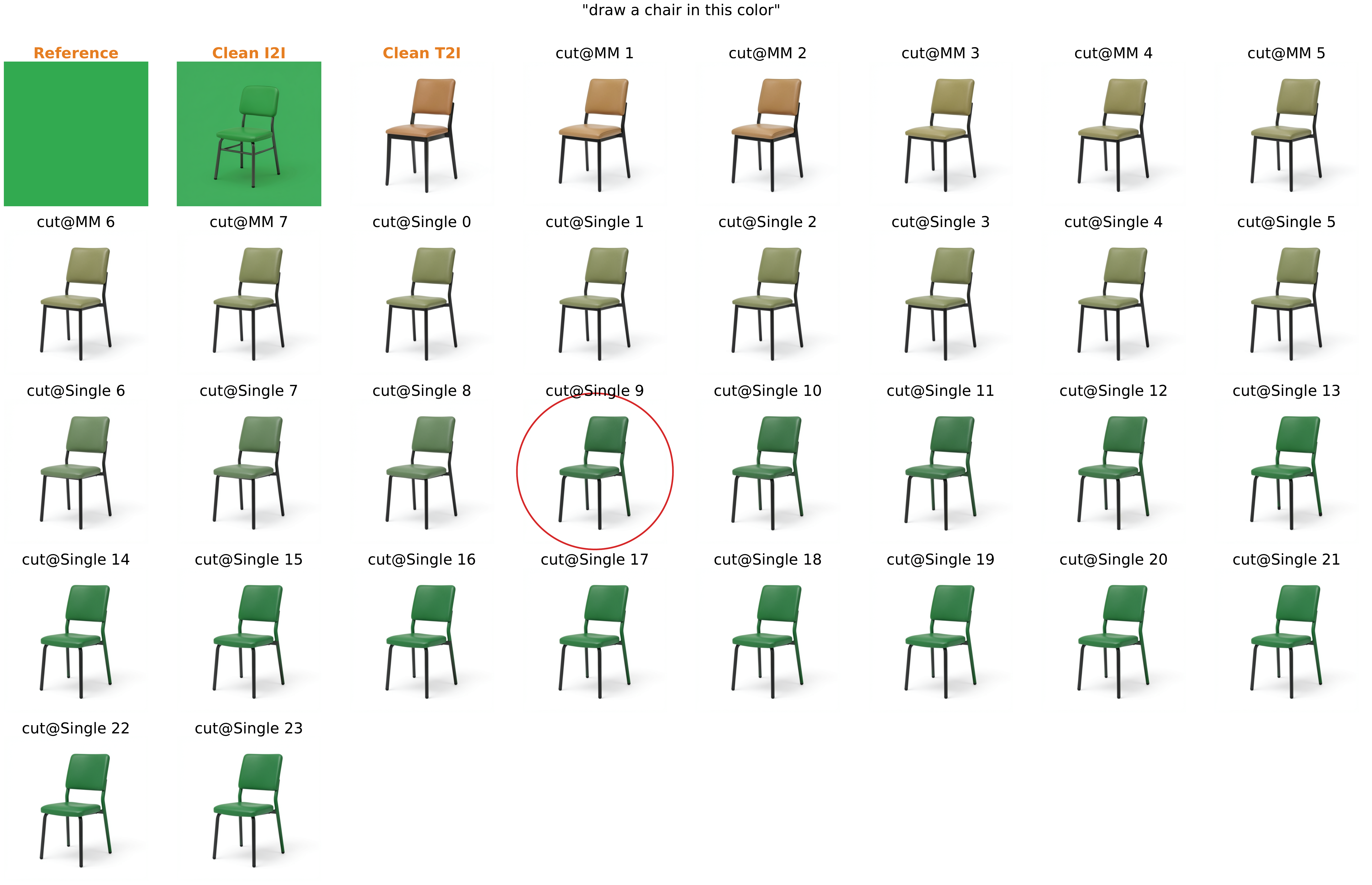}
    \caption{\textbf{Sweeping the reference-drop cutoff on a color transfer task.} Each
    cell runs the experiment of Figure~\ref{fig:early_ref_drop} on a color edit (drawing a
    green chair), with the cutoff moved to a progressively later block. The reference green
    is absent when the cutoff is early and is fully present once the cutoff passes the
    10th single stream block, matching the color binding layer in
    Appendix~\ref{subsec:t2i_color}.}
    \label{fig:split_grid_color}
\end{figure}

\clearpage
\clearpage

\section{More Qualitative Results}

\begin{figure}[h]
    \centering
    \includegraphics[width=1\linewidth]{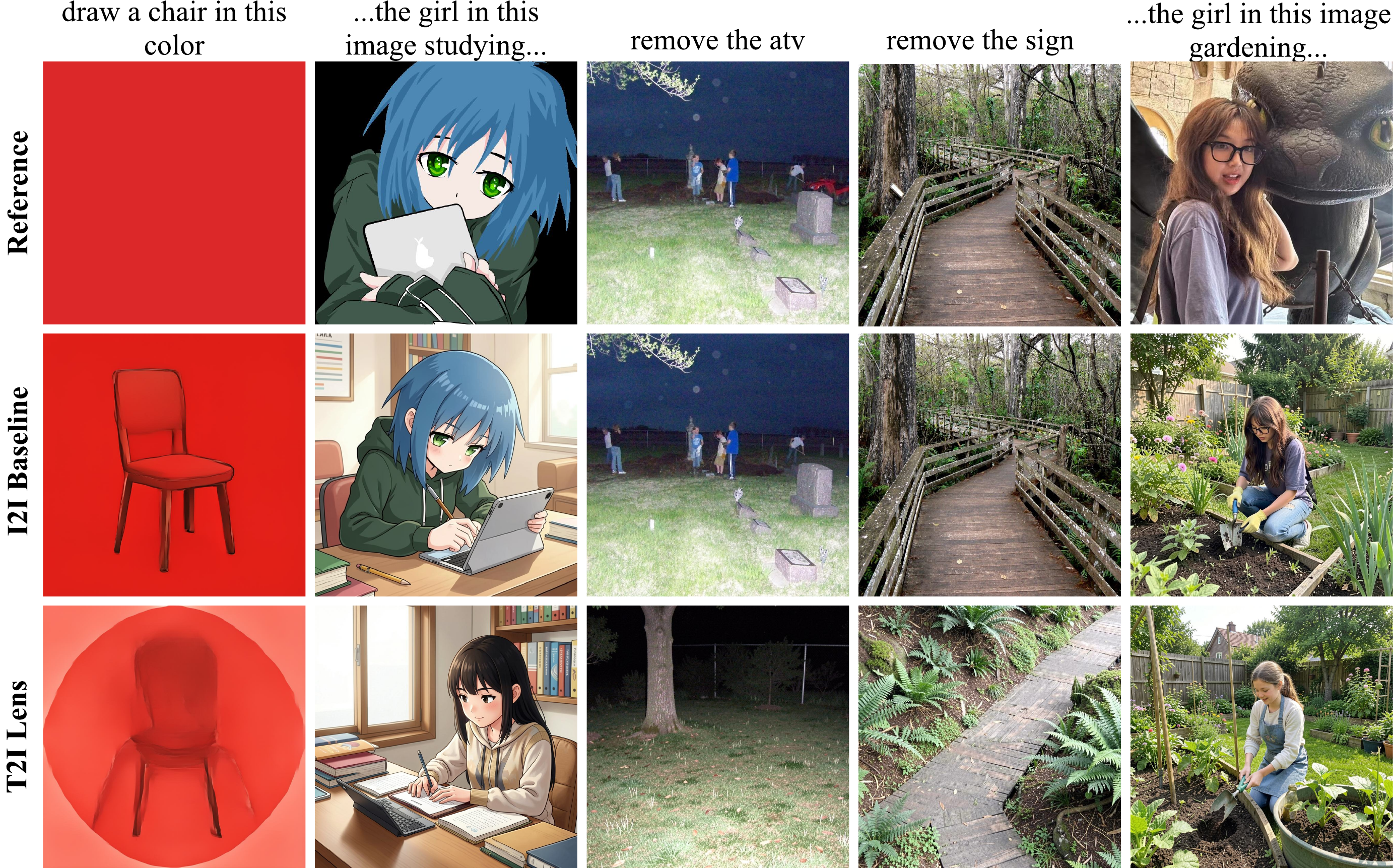}
    \caption{\textbf{More T2I Lens examples}. We include color and style examples that were not present in the main paper, demonstrating that the reference color and unrealism are conveyed by the text tokens. Recall from Appendix \ref{subsec:t2i_color} that the color-specific T2I Lens is a single step, which is responsible for its poor image quality. As in Section \ref{subsec:t2i}, we see that neither the identity of the real girl nor the fictional girl is preserved in the text tokens (even though whether the result is real or fictional is preserved). }
    \label{fig:more_t2i_lens}
\end{figure}

\begin{figure}[h]
    \centering
    \includegraphics[width=1\linewidth]{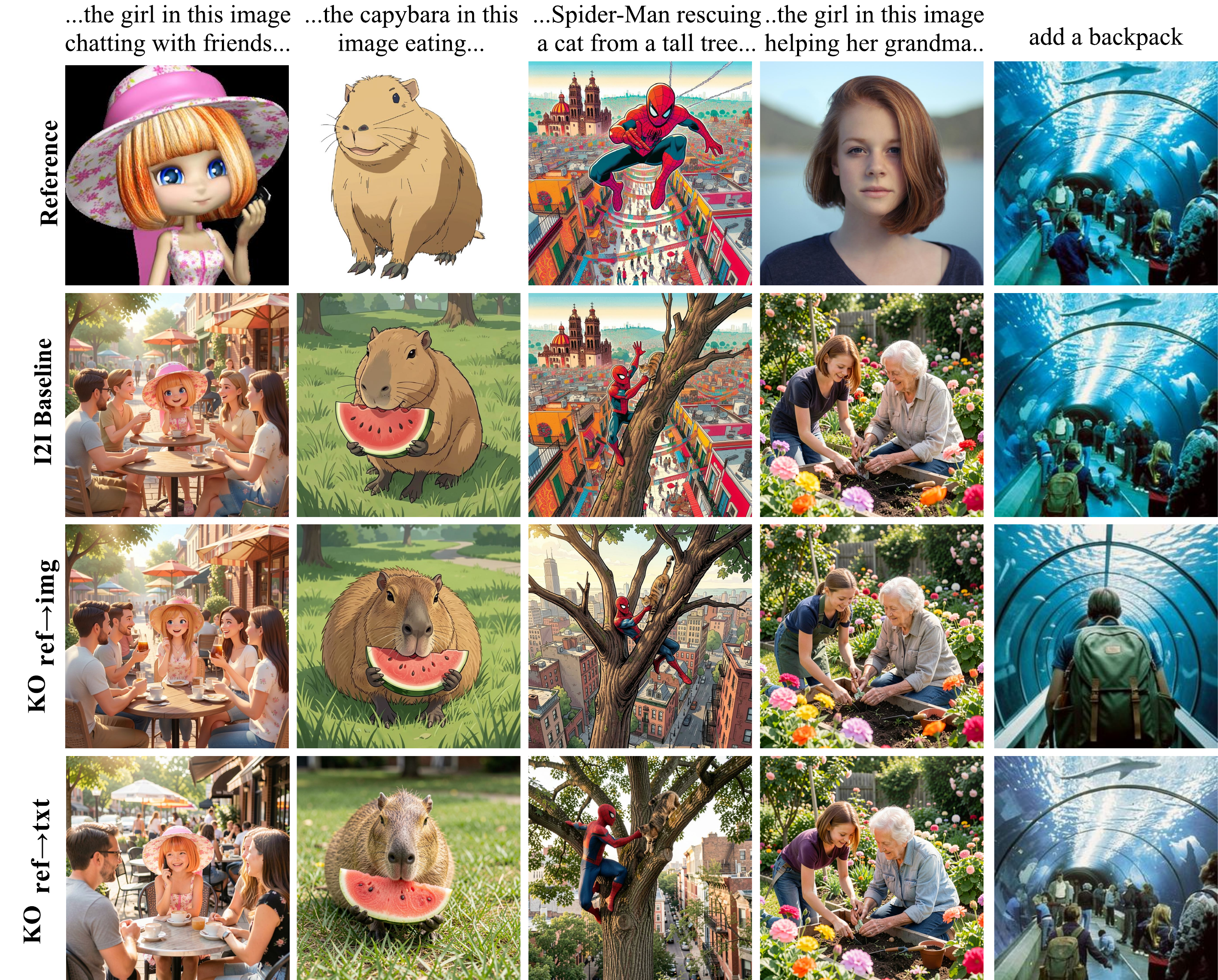}
    \caption{\textbf{More attention knockout examples.} The first three columns follow the same trend where knocking out ref->text results in the stylized image becoming more realistic. In the fourth column, we see that knocking out ref->image has a larger effect on the human identity, though the girl did get younger and have a different shirt color when knocking out ref->text. In the last column, knocking out ref->text preserves all details besides the saturation of the image, furthering the claim that color is one of the few targeted roles of the text tokens' reference information. }
    \label{fig:more_attention_knockout}
\end{figure}

\begin{figure}[H]
    \centering
    \includegraphics[width=1\linewidth]{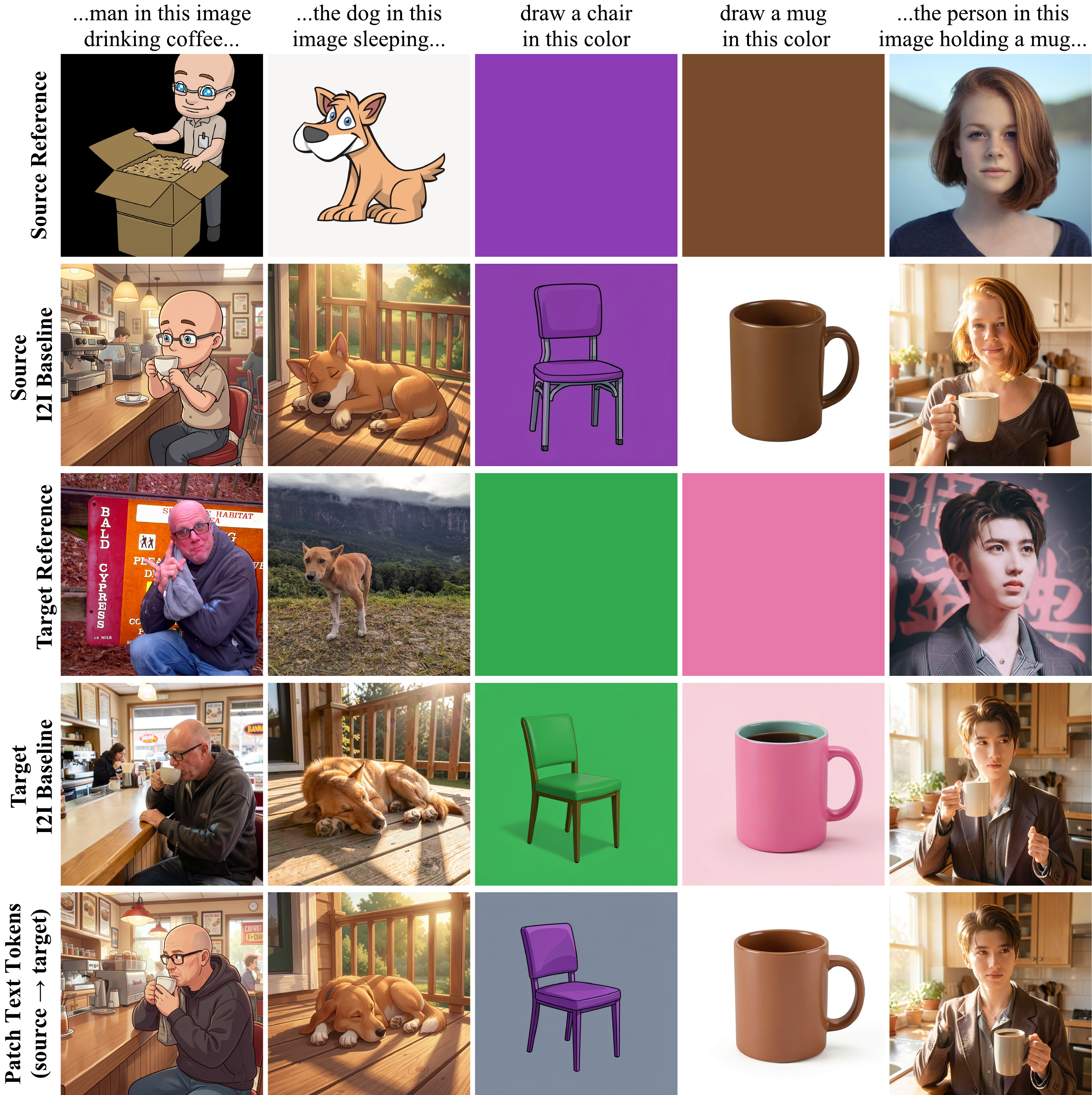}
    \caption{\textbf{More I2I-to-I2I Patching Examples.} These examples also show transferring style (first two columns) and color (columns three and four)  without affecting other details and failure to transfer human identity (column five). }
\label{fig:more_i2i_to_i2i_patching}
\end{figure}

\begin{figure}[H]
    \centering
    \includegraphics[width=1.0\linewidth]{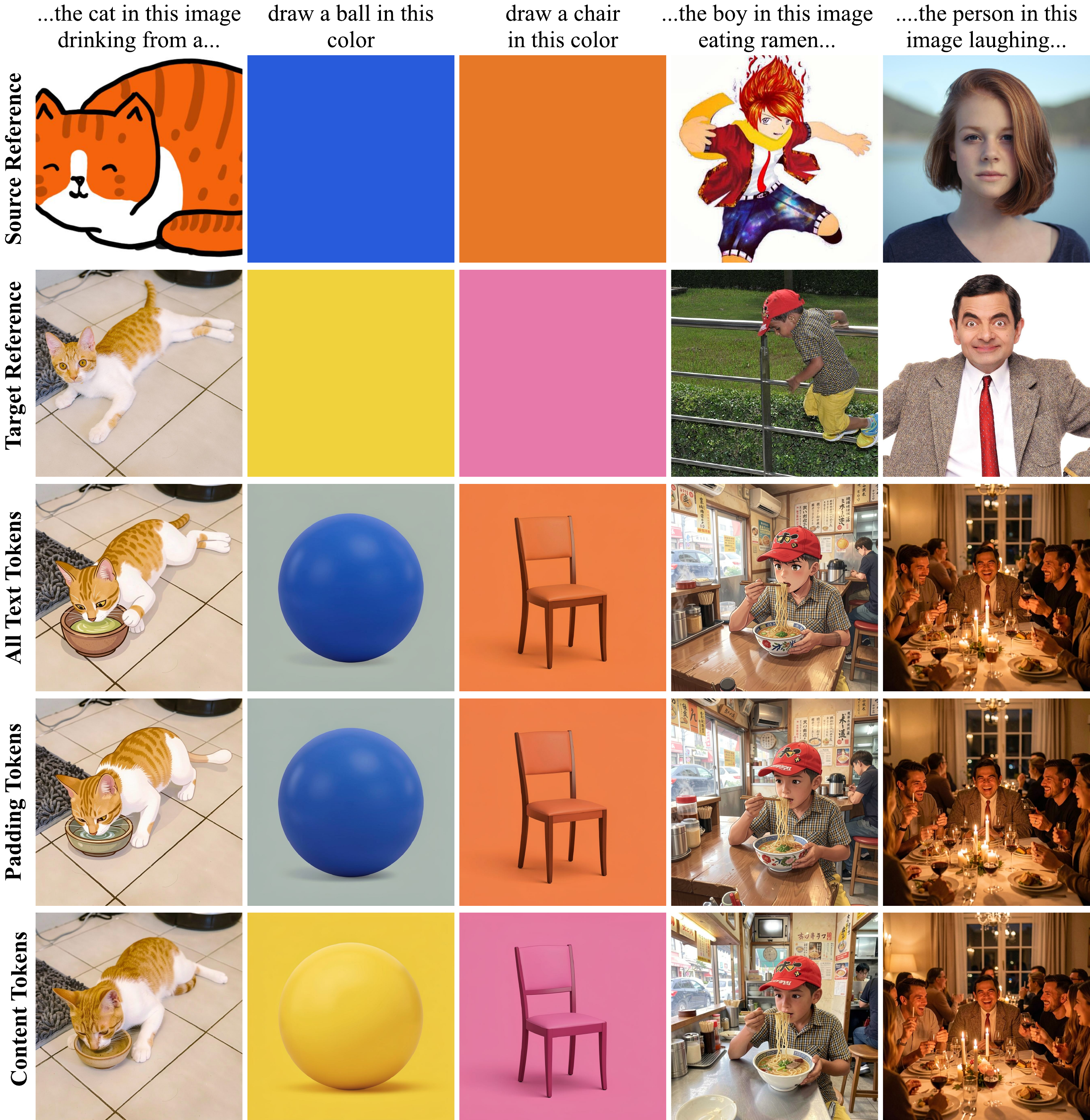}
    \caption{\textbf{I2I-to-I2I Patching with just padding or just content text tokens}. We redo the I2I-to-I2I patching experiments but patching only the padding text tokens or patching only the content text tokens. We find that patching just the padding tokens successfully transfers the color completely (columns two and three), and almost completely transfers the style (columns 1 and 4), leaving just a little hint of realism. In contrast, patching only the content tokens does not transfer color or style. None of the approaches have an effect on transferring human identity (column five). }
    \label{fig:i2i_patching_padding}
\end{figure}

Here we provide more qualitative results for the three experiments from the main paper in Figures \ref{fig:more_t2i_lens}, \ref{fig:more_attention_knockout}, \ref{fig:more_i2i_to_i2i_patching}, in the same format of Figures \ref{fig:T2I_lens_qual}, \ref{fig:attention_knockout}, \ref{fig:i2i_to_i2i}. We also include the I2I-to-I2I patching results for only padding/content text tokens in Figure \ref{fig:i2i_patching_padding}.

\subsection{T2I Lens}
\label{subsec:more_T2I_lens}

\clearpage

\subsection{Attention Knockout} 
\label{subsec:more_attention_knockout}
\clearpage

\subsection{I2I-to-I2I Patching}
\label{subsec:more_i2i_to_i2i_patching}

\subsubsection{I2I-to-I2I Patching using only padding tokens or only content tokens}

\clearpage

\section{Full Attention Knockout Table}
\label{sec:attention-knockout-full}

For attention knockout, we find similar trends in the results to the I2I-to-I2I patching case in that knocking out attention to just the padding tokens still often results in a removal of the reference's color or style, while knocking out attention to just the content tokens does not cause the same effect. However, for attention knockout, the results are less clear-cut than the I2I-to-I2I patching case. This may be due to the fact that knocking out attention to only the padding tokens may reroute information to the content tokens instead, or invoke some kind of backup mechanism. 

\begin{table}[H]
\centering
\caption{Full attention knockout sweep, including the padding-only and content-only $\text{KO}_{\text{ref}\rightarrow\text{text}}$ variants left out of the main-body table.}
\label{tab:attention_knockout_full}
% AUTO-TABLE: attention_knockout_full START — generated by scripts/build_judge_tables.py
\begin{tabularx}{\linewidth}{l >{\centering\arraybackslash}X >{\centering\arraybackslash}X >{\centering\arraybackslash}X}
\toprule
 & Color Transfer (\%) & Style Transfer (\%) & Human Customization (\%) \\
\midrule
KO\textsubscript{ref$\rightarrow$text} & $86.2_{-4.2}^{+3.3}$ & $97.3_{-1.9}^{+1.1}$ & $45.6_{-9.9}^{+10.3}$ \\
KO\textsubscript{ref$\rightarrow$text[padding]} & $70.0_{-5.2}^{+4.8}$ & $62.9_{-4.6}^{+4.3}$ & $16.7_{-6.3}^{+9.0}$ \\
KO\textsubscript{ref$\rightarrow$text[content]} & $5.0_{-1.9}^{+3.0}$ & $27.6_{-3.9}^{+4.3}$ & $7.8_{-4.0}^{+7.4}$ \\
KO\textsubscript{ref$\rightarrow$image} & $13.4_{-3.3}^{+4.2}$ & $16.4_{-3.1}^{+3.7}$ & $76.7_{-9.7}^{+7.5}$ \\
\bottomrule
\end{tabularx}
% AUTO-TABLE: attention_knockout_full END
\end{table}

\section{VLM-as-judge prompts}
\label{sec:vlm-judge-prompts}

Every quantitative metric reported is the mean of binary verdicts produced by Claude Opus 4.7 acting as a visual judge. Each judge call includes a fixed system prompt, a sequence of 2--4 labeled images, and one closed-form question; the model answers with a single-line JSON object whose \texttt{pass} field is 1 when the prediction holds and 0 otherwise. The system prompt below is shared across every judge in the paper.

\begin{promptbox}[title=Judge system prompt]
You are a strict visual judge for an image-editing interpretability experiment. You will be shown several labeled images and asked one yes/no question about whether a stated prediction is satisfied.
Respond ONLY with a single JSON object on one line, no markdown, no preamble:
{"pass": 0 or 1, "reason": "<one short sentence, <=25 words>"}
The "pass" field is 1 if the prediction is satisfied, 0 otherwise. If the question cannot be answered from the images, reply with {"pass": 0, "reason": "cannot determine"}.
\end{promptbox}

\subsection{T2I Lens}
\label{subsec:judge-t2i-lens}

The T2I Lens judges receive three images: the reference, a T2I baseline using the same prompt as the I2I, and the patched unconditional T2I. The question asks whether information about the reference appears in the patched generation (despite the reference never directly conditioning that pass). For the color transfer tasks, it asks if the color appears in the patched image; for styled reference image tasks, it asks if the style of the reference image appears in the patched image; for human customization tasks, it asks if the identity of the human fails to appear in the patched image; and for object addition / removal tasks, it asks if any visual aspects of the reference setting appear in the patched image. 

\begin{promptbox}[title=Color Transfer]
Compared to Image 2, does Image 3 take on the predominant solid color of Image 1? Reply 1 if the color of Image 1 is now visibly present in Image 3 (and was not in Image 2). 
\end{promptbox}

\begin{promptbox}[title=Style Transfer]
Compared to Image 2, does Image 3 adopt a clipart / cartoon / illustrated / unrealistic style similar to Image 1? Look at the subject and the background / rest of the image - clipart-y style anywhere in the image counts as evidence. Reply 1 if Image 3 looks more clipart-like / less photographic than Image 2.
\end{promptbox}

\begin{promptbox}[title=Human Customization Identity]
Focus on the person in Image 3. Is the person in Image 3 a recognizably DIFFERENT individual (different face, hair, build, identity) from the person in Image 1? Reply 1 if a viewer would say it is a different person; reply 0 only if it is the same person.
\end{promptbox}

\begin{promptbox}[title=Object Addition and Removal from Scene]
Compared to Image 2, does Image 3 contain ANY visible information drawn from Image 1 - things like colors, textures, layout, style, distinctive shapes, or specific subject features? Reply 1 if you can point to anything in Image 3 that came from Image 1 and was not in Image 2.
\end{promptbox}

\subsection{Attention Knockout}
\label{subsec:judge-attention-knockout}

The attention knockout judges receive three images: the reference, the reference-conditioned I2I baseline, and the same generation with one attention path blocked. We split the prompts by the property tested (color, style, human identity). Each property has two judges, one for the \texttt{ref->text} direction and one for the \texttt{ref->image} direction. The two share the same images but flip polarity: the \texttt{ref->text} judge asks whether the property has been lost when the attention is blocked, while the \texttt{ref->image} judge asks whether the property is still preserved when the attention is blocked. Within \texttt{ref->text}, the three variants for knocking out all text tokens, only padding text tokens, or only content text tokens reuse the same question with the parenthetical \texttt{ref->text} replaced by \texttt{ref->text[padding]} or \texttt{ref->text[content]}; we show only the full-variant wording.

\begin{promptbox}[title=Color (solid\_color) -- ref->text]
Compared to Image 2, has Image 3 LOST the predominant solid color of Image 1? Reply 1 if the color is significantly removed (color depended on ref->text).
\end{promptbox}

\begin{promptbox}[title=Clipart style -- ref->text]
Compared to Image 2, has Image 3 LOST the clipart / cartoon style of Image 1 and become more photographic / realistic? Reply 1 if Image 3 became more realistic when ref->text was blocked.
\end{promptbox}

\begin{promptbox}[title=DreamBench++ humans (identity) -- ref->text]
Focus on the person in Image 3. Compared to Image 2, has Image 3 LOST the identity of the person in Image 1 - i.e. does the person in Image 3 look like a recognizably DIFFERENT individual (different face, hair, build) from the person in Image 1? Reply 1 if blocking ref->text destroyed the reference identity; reply 0 if Image 3 still looks like the same person as Image 1.
\end{promptbox}

\subsection{I2I-to-I2I Patching}
\label{subsec:judge-i2i-i2i}

The I2I-to-I2I judges receive four images: the source reference, the target reference, the target I2I baseline, and the patched target generation. We test whether the source property transfers over to the target. The full / padding-only / content-only variants share the same question.

\begin{promptbox}[title=Color pair]
Compared to Image 3 (which should show the color of Image 2), does Image 4 take on the color of Image 1 (the source) instead? Reply 1 if Image 4 is more like Image 1's color than Image 2's.
\end{promptbox}

\begin{promptbox}[title=Clipart pair (clipart source -> real-photo target)]
Compared to Image 3, has Image 4 become MORE clipart / cartoon / unrealistic in style (matching Image 1)? Look at the subject and the background / rest of the image - clipart-y style anywhere in the image counts as evidence. Reply 1 if Image 4 looks more clipart-like than Image 3.
\end{promptbox}

\begin{promptbox}[title=DreamBench++ humans pair]
Focus on the person in Image 4. Does the person in Image 4 look more like person A (Image 1, the source) than like person B (Image 2, the target)? Reply 1 if A's identity transferred over.
\end{promptbox}

\section{VLM-assisted instruction generation}
\label{sec:vlm-instruction-generation}

Four of the task families (object addition, object removal, human customization, style transfer) either fully delegate or hybridize the task instruction-writing step to a VLM API call (Claude Opus 4.7). We present the prompts used to generate diverse, appropriate task instructions that don't reveal specific information about the reference image below. 

\subsection{Object addition / Object removal (SUN397)}
\label{subsec:vlm-sun397}

For every SUN397 image we make a single VLM call that returns two short noun phrases (one each for add and remove), which we then template into ``add the \emph{<noun>}'' and ``remove the \emph{<noun>}'' instructions.

\begin{promptbox}[title=SUN397 instruction-generation prompt]
You design test cases for an image-editing experiment. Given ONE image,
propose two short object names:

1. ADD: a single object NOT currently in the image but plausibly fits
   the scene. If no scene-agnostic addition fits, set "add_object": null.
2. REMOVE: a single object IS visible and could be plausibly removed
   (not the entire subject; the scene would still read coherently
   without it). If nothing meets that bar, set "remove_object": null.

SCENE-AGNOSTIC RULE: the proposed object names must NOT reveal the
specific scene/location depicted. For a volcano photo, "lava plume" or
"volcanic crater" is forbidden -- those give away the scene. Prefer
generic objects that could plausibly fit many different scenes. If no
scene-agnostic object fits a given field, set it to null.

FORBIDDEN OBJECTS: do NOT propose any of these (or any phrase
containing one of these words as the head noun): bird, backpack,
person, bicycle, bench, bottle, plant, dog. They are over-represented
in our existing dataset. Pick a different scene-agnostic object, or
set the field to null if nothing else fits.

Return ONLY a single JSON object on one line, no markdown:
{"add_object": "<noun>" | null,
 "remove_object": "<noun>" | null}
 
Rules:
- Object phrases: 1-4 words, lowercase, no punctuation, singular.
- All non-null object names must be pairwise distinct.
- No proper nouns, no named people.
- All non-null object names must obey the SCENE-AGNOSTIC RULE.
- All non-null object names must NOT be in the FORBIDDEN OBJECTS list.
\end{promptbox}

\subsection{Style Transfer Tasks}
\label{subsec:style_transfer_instructions}

For each of the 18 manually chosen fictional image subjects, we first hand-wrote one instruction following the template ``A photograph of the \emph{<subject>} in this image \emph{<verb-phrase>}''. We then prompt Claude Opus 4.7 for four additional instructions in the same template, yielding 5 prompts per subject (90 instructions total). The system prompt and user template are below; \texttt{\{subject\}}, \texttt{\{slug\}}, and \texttt{\{existing\_prompt\}} are substituted per call.

\begin{promptbox}[title=Style transfer instruction system prompt]
You write short action/scene prompts for image-customization tasks. Each prompt MUST follow the exact template: "A photograph of the {subject} in this image {verb_phrase}." The verb phrase should describe a single concrete action, scene, or interaction (5 to 15 words). Make each prompt clearly distinct from the others and from the example. No metaphor, no abstract concepts, no body-part close-ups.
\end{promptbox}

\begin{promptbox}[title=Style transfer instruction user template]
Subject: {subject}
Existing prompt (slug={slug}): {existing_prompt}

Write 4 NEW prompts following the same template, each describing a different action or scene from the existing one. Output them as a JSON array of 4 objects, each with fields "slug" (a short snake_case verb-phrase identifier, e.g. "playing_chess") and "prompt" (the full sentence). No prose around the JSON.
\end{promptbox}

\subsection{Human Customization Tasks (DreamBench++)}
\label{subsec:vlm-dreambench-humans}
For the human customization tasks, in addition to the individualized prompts per human subject, we require a shared small set of generic, person-agnostic instructions, so that for the I2I-to-I2I patching experiments, we can pair different humans without leaking subject-specific cues, while still having the same source and target edit instructions. We generate these with the same VLM prompts as Section~\ref{subsec:style_transfer_instructions} (with \texttt{\{subject\}} = ``person''). The five final instructions, hand-curated from the VLM output, are:

\begin{itemize}
  \item ``A photograph of the person in this image walking down a busy city street at golden hour''
  \item ``A photograph of the person in this image holding a mug of coffee in a sunlit kitchen''
  \item ``A photograph of the person in this image sitting on a wooden park bench reading a book''
  \item ``A photograph of the person in this image standing at the edge of a foggy beach at dawn''
  \item ``A photograph of the person in this image laughing at a candlelit dinner party with friends''
\end{itemize}

\section{Compute budget and asset licenses}
\label{sec:compute_and_licenses}

\paragraph{Compute budget.}
We run experiments on a shared cluster on a variety of GPU types, including
H200, H100, A100, L40S, RTX A6000, and RTX 6000 Ada GPUs. Running all
experiments reported in the paper took approximately 18 H200 GPU-hours;
preliminary and discarded experiments add a small amount on top of that.

\paragraph{Asset licenses.}
We list every external asset used to produce the results, with version, URL, and license.

\textit{Datasets.}
\begin{itemize}
    \item \textbf{SUN397} \citep{xiao2010sun}, \url{https://vision.princeton.edu/projects/2010/SUN/}.
          We use 2 images per category across all 397 scene categories (794 images total)
          as references for the \emph{add} (789 tasks) and \emph{remove} (726 tasks) buckets.
          SUN397 does not publish an explicit dataset-level license; the original photos were
          web-crawled and each retains its photographer's copyright. Our use is non-commercial
          academic research.
    \item \textbf{DreamBench++} \citep{peng2025dreambench++},
          \url{https://huggingface.co/datasets/yuangpeng/dreambench_plus} (Apache-2.0).
          We use 10 real-human reference images from the \texttt{live\_subject/human}
          category (140 customize tasks). The dataset's authors note that images originate
          from their own collection, Unsplash, Rawpixel, and Google Image Search, with
          copyright verified for academic suitability.
\end{itemize}

\textit{Reference images (style transfer).}
The 450 style-transfer customize tasks use 18 paired (illustration, real-photo)
reference images, each sourced under an open license. Several assets carry
CC BY-NC (the iNaturalist photos under 4.0, one Openverse illustration under
2.0), which permits our non-commercial academic use; the remainder are
public-domain works or permissive Creative Commons / CC0.

\begin{itemize}
    \item \textbf{alice\_in\_wonderland} — illustration:
          \url{https://commons.wikimedia.org/wiki/File:De_Alice%27s_Abenteuer_im_Wunderland_Carroll_pic_04.jpg}
          (public domain, 1869). Real photo:
          \url{https://commons.wikimedia.org/wiki/File:Little_girl_in_a_Kazakh_national_dress.jpg}
          (CC BY-SA 4.0).
    \item \textbf{anime\_girl} — illustration:
          \url{https://openclipart.org/detail/250032/girl-with-a-tablet} (CC0). Real photo:
          \url{https://commons.wikimedia.org/wiki/File:Asian_girl_with_dimples.jpg}
          (CC BY 2.0).
    \item \textbf{bald\_man} — illustration:
          \url{https://commons.wikimedia.org/wiki/File:Bald_Man_Opening_A_Package_Cartoon.svg}
          (CC BY-SA 4.0). Real photo:
          \url{https://commons.wikimedia.org/wiki/File:Bald_man_bald_cypress_with_my_best_friend_(23884877910).jpg}
          (CC BY 2.0).
    \item \textbf{boy\_pixel} — illustration:
          \url{https://openverse.org/image/7e411861-5f6c-41b6-b1d2-ddf5569241fb}
          (CC0 1.0). Real photo:
          \url{https://commons.wikimedia.org/wiki/File:Boy_with_Dog_-_Parque_Mexico_-_Condesa_District_-_Mexico_City_-_Mexico_(15333791897).jpg}
          (CC BY-SA 2.0).
    \item \textbf{capybara} — illustration:
          \url{https://openclipart.org/detail/348465/capybara} (CC0). Real photo:
          \url{https://www.inaturalist.org/photos/91546944} (CC BY-NC 4.0).
    \item \textbf{cartoon\_boy} — illustration:
          \url{https://openverse.org/image/adbc82f6-6de0-4c2c-8ea2-091a005068f0}
          (CC BY 2.0). Real photo:
          \url{https://commons.wikimedia.org/wiki/File:Teenage_redhead_boy_proudly_holding_fresh_caught_bass.jpg}
          (public domain).
    \item \textbf{dog} — illustration:
          \url{https://openverse.org/image/6ad00da9-eedd-454b-875b-eff32ac40410}
          (CC0 1.0). Real photo:
          \url{https://www.inaturalist.org/observations/173164942} (CC BY-NC 4.0).
    \item \textbf{doll} — illustration:
          \url{https://openclipart.org/detail/245209/fashionable-cartoon-girl} (CC0). Real
          photo: \url{https://commons.wikimedia.org/wiki/File:Lucy_Merriam.jpg}
          (CC BY-SA 3.0).
    \item \textbf{elephant} — illustration:
          \url{https://commons.wikimedia.org/wiki/File:I_AM_SAMBHU.jpg}
          (CC BY-SA 4.0). Real photo:
          \url{https://www.inaturalist.org/photos/113246959} (CC BY-NC 4.0).
    \item \textbf{fox} — illustration:
          \url{https://openverse.org/image/a6076556-88ca-4881-9770-5e7ad6a3a59c}
          (CC BY-NC 2.0). Real photo:
          \url{https://www.inaturalist.org/observations/5305215} (CC BY-NC 4.0).
    \item \textbf{frog} — illustration:
          \url{https://openverse.org/image/0f185ddd-9807-4497-b125-2b40ef02fd8b}
          (CC0 1.0). Real photo:
          \url{https://www.inaturalist.org/observations/248834388} (CC BY-NC 4.0).
    \item \textbf{globe\_man} — illustration:
          \url{https://openclipart.org/detail/331082/fat-america-man-colour-remix} (CC0).
          Real photo: \url{https://commons.wikimedia.org/wiki/File:Man_in_black_suit_and_white_shirt_with_hands_in_pockets,_looking_extremely_badass_cool.jpg}
          (CC BY-SA 4.0).
    \item \textbf{mouse} — illustration:
          \url{https://openverse.org/image/ecb50fd5-4d28-4692-a594-65bc439b5694}
          (CC0 1.0). Real photo:
          \url{https://www.inaturalist.org/observations/35194886} (CC BY-NC 4.0).
    \item \textbf{orange\_cat} — illustration:
          \url{https://commons.wikimedia.org/wiki/File:Orange_cat_cartoon.png}
          (CC BY-SA 4.0). Real photo:
          \url{https://commons.wikimedia.org/wiki/File:Melon,_the_orange_cat,_Taipei;_July_2016_(17).jpg}
          (CC BY-SA 2.0).
    \item \textbf{pig} — illustration:
          \url{https://commons.wikimedia.org/wiki/File:Pig_cartoon_04.svg} (public domain).
          Real photo: \url{https://commons.wikimedia.org/wiki/File:Cochon_domestique_(Sus_scrofa_domesticus)_(2).jpg}
          (CC BY-SA 4.0).
    \item \textbf{solarmax} — illustration:
          \url{https://openverse.org/image/fd293c75-e1f2-4f36-b8e4-9ba28ae6258a}
          (CC BY 2.0). Real photo:
          \url{https://commons.wikimedia.org/wiki/File:Boy_Jumping_Over_Railings_-_Science_City_-_Kolkata_2018-09-23_4270.JPG}
          (CC BY-SA 4.0).
    \item \textbf{sparrow} — illustration:
          \url{https://openverse.org/image/a0fb7533-15d4-4300-89fe-0a1de20644a3}
          (CC0 1.0). Real photo:
          \url{https://www.inaturalist.org/observations/1566579} (CC BY-NC 4.0).
    \item \textbf{woman\_runner} — illustration:
          \url{https://openclipart.org/detail/269990/cartoon-woman-portrait} (public domain).
          Real photo: \url{https://commons.wikimedia.org/wiki/File:Smiling_Woman_Runner_(8425517545).jpg}
          (CC BY-SA 2.0).
\end{itemize}

\textit{Model and tokenizer.}
\begin{itemize}
    \item \textbf{FLUX.2-klein-9B} (Black Forest Labs),
          \url{https://huggingface.co/black-forest-labs/FLUX.2-klein-9B}, released under
          the FLUX Non-Commercial License v2.0
          (\url{https://bfl.ai/legal/non-commercial-license-terms}). We use the released
          weights at inference only; we do not train, fine-tune, or distill, and our use is
          non-commercial academic research.
    \item \textbf{Qwen3-Embedding-8B} (Alibaba), Apache-2.0;
          \url{https://huggingface.co/Qwen/Qwen3-Embedding-8B}. Used as the text encoder
          inside \texttt{Flux2KleinPipeline}.
\end{itemize}

\textit{APIs.}
\begin{itemize}
    \item \textbf{Anthropic Claude API.} We use Claude Opus 4.7 to generate short
          noun-phrase add/remove instructions for the 794 SUN397 references and as the
          VLM-as-judge for all results CSVs (Appendix~\ref{sec:vlm-instruction-generation}). Calls go
          through Anthropic's commercial API under the Anthropic Usage Policy and API
          Terms of Service (\url{https://www.anthropic.com/legal}). We redistribute
          neither the model weights nor a derived dataset of generations.
\end{itemize}

%%%%%%%%%%%%%%%%%%%%%%%%%%%%%%%%%%%%%%%%%%%%%%%%%%%%%%%%%%%%

% \newpage
% \input{checklist.tex}

\end{document}